 \documentclass[pmlr,twocolumn,10pt]{template/jmlr} 

\usepackage{booktabs}
\usepackage{siunitx}

\newcommand{\equal}[1]{{\hypersetup{linkcolor=black}\thanks{#1}}}

\theorembodyfont{\upshape}
\theoremheaderfont{\scshape}
\theorempostheader{:}
\theoremsep{\newline}

\jmlrvolume{225}
\jmlryear{2023}
\jmlrsubmitted{LEAVE UNSET}
\jmlrpublished{LEAVE UNSET}
\jmlrworkshop{Machine Learning for Health (ML4H) 2023} 

\usepackage[acronym]{glossaries}
\newacronym{ai}{AI}{artificial intelligence}
\newacronym{ml}{ML}{machine learning}
\newacronym{dl}{DL}{deep learning}
\newacronym{dcai}{DCAI}{data-centric artificial intelligence}
\newacronym{dino}{DINO}{self-distillation with no labels}
\newacronym{isic}{ISIC}{International Skin Imaging Collaboration}
\newacronym{vit}{ViT}{vision transformer}
\newacronym{vit-t}{ViT-T}{ViT-Tiny}
\newacronym{ssl}{SSL}{self-supervised learning}
\newacronym{simclr}{SimCLR}{simple framework for contrastive learning of visual representations}
\newacronym{knn}{kNN}{$k$-nearest neighbor}
\newacronym{auroc}{AUROC}{area under the ROC curve}
\newacronym{ap}{AP}{average precision}
\newacronym{auprg}{AUPRG}{area under the precision-recall-gain curve}
\glsunset{simclr}
\glsunset{dino}

 \title[Towards Reliable Dermatology Evaluation Benchmarks]{Towards Reliable Dermatology Evaluation Benchmarks}
 
\author{%
\Name{Fabian Gr\"oger}$^{1,2}$ \Email{fabian.groeger@unibas.ch} \\
\Name{Simone Lionetti}$^2$ \Email{simone.lionetti@hslu.ch} \\
\Name{Philippe Gottfrois}$^1$ \Email{philippe.gottfrois@usb.ch} \\
\Name{Alvaro Gonzalez-Jimenez}$^1$ \Email{alvaro.gonzalezjimenez@unibas.ch} \\
\Name{Matthew Groh}$^3$ \Email{matthew.groh@kellogg.northwestern.edu} \\
\Name{Roxana Daneshjou}$^4$ \Email{roxanad@stanford.edu} \\
\Name{Labelling Consortium}$^{2,5,6,}$\equal{Florian B\"ar, Pascal Baumann, Anna-Maria Forster, Elisabeth G\"ossinger, Antonios Kolios, Ramon Koller, and Alina M\"uller} \Email{dermanatomy@usb.ch} \\
\Name{Alexander A.\ Navarini}$^{1,5}$ \Email{alexander.navarini@usb.ch} \\
\Name{Marc Pouly}$^2$ \Email{marc.pouly@hslu.ch} \\
\addr 
$^1$Department of Biomedical Engineering, University of Basel, Switzerland \\
$^2$Department of Computer Science, Lucerne University of Applied Sciences and Arts, Switzerland \\
$^3$Kellogg School of Management, Northwestern University, USA \\
$^4$Department of Dermatology, Stanford School of Medicine, USA \\
$^5$Department of Dermatology, University Hospital of Basel, Switzerland \\
$^6$Department of Dermatology, University Hospital of Zurich, Switzerland 
}

\begin{document}
\frenchspacing

\maketitle

\begin{abstract}
Benchmark datasets for digital dermatology unwittingly contain inaccuracies that reduce trust in model performance estimates.
We propose a resource-efficient data-cleaning protocol to identify issues that escaped previous curation.
The protocol leverages an existing algorithmic cleaning strategy and is followed by a confirmation process terminated by an intuitive stopping criterion.
Based on confirmation by multiple dermatologists, we remove irrelevant samples and near duplicates and estimate the percentage of label errors in six dermatology image datasets for model evaluation promoted by the \acrlong*{isic}.
Along with this paper, we publish revised file lists for each dataset which should be used for model evaluation.\footnote{\url{https://github.com/Digital-Dermatology/SelfClean-Revised-Benchmarks}}
Our work paves the way for more trustworthy performance assessment in digital dermatology. 
\end{abstract}

\begin{figure*}[t]
\floatconts
  {fig:Teaser}
  {\caption{
      Examples of data quality issues found in the six considered evaluation datasets for dermatology.
      PH2 is not shown, as it was found to contain no issues.
  }}
  {\includegraphics[width=1.0\linewidth]{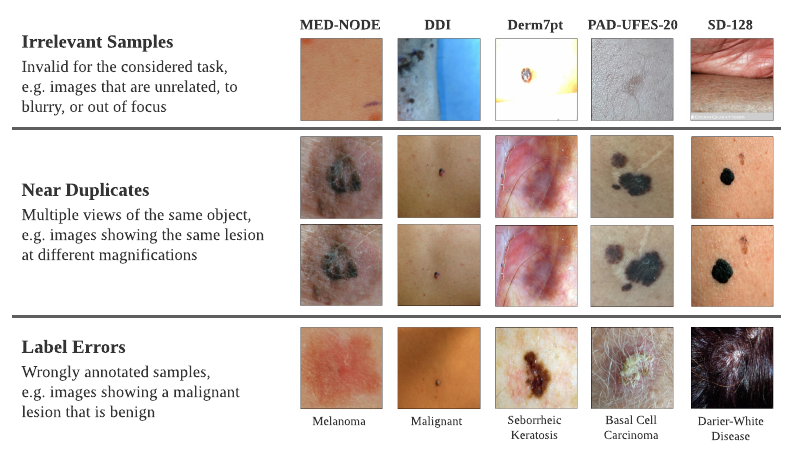}}
\end{figure*}

\section{Introduction}
Benchmark datasets serve as proxies to evaluate the utility of \gls*{ml} models in real-life tasks, e.g. to estimate clinical utility in the medical domain.
In particular, the scientific community actively promotes the creation, distribution, and adoption of public benchmark datasets to enhance comparability and reproducibility.
New benchmark datasets are typically introduced by a work that observes their specific merits or by a competition that addresses the lack of satisfactory solutions for a practically relevant problem.
After their introduction, however, they are often used without further critical investigation.
While this fulfills the purpose of fair comparison among different methods, the degree to which performance reflects real utility remains an open question.
Indeed, concerns have been raised regarding the quality of existing benchmarks~\citep{groh2022identifying}.
For example, \citet{northcutt_pervasive_2021} identified label errors across domains and argued that noise in benchmark datasets can obscure the true model performance.
Consequently, questions arise regarding the validity of leaderboards and the effectiveness of increasingly sophisticated approaches.

Dermatology is no exception in this respect.
For instance, \citet{tschandl_humancomputer_2020,tschandl_domain-specific_2019} found the amount of noise present in annotations to be a limiting factor, which prompted the authors to repeat the manual segmentation process.
Furthermore, some popular dermatology datasets such as Fitzpatrick17k \citep{groh_evaluating_2021} and SD-128/192/256 \citep{leibe_benchmark_2016} have been obtained by crawling atlases or collecting user inputs.
These strategies are inherently prone to errors, which leads to relatively high noise levels \citep{sambasivan_everyone_2021}.
As an alternative for more reliable evaluation, there has been a surge in the creation of small datasets designed for evaluation purposes.
These datasets typically prioritize the elimination of biases, aim to represent data diversity, and undergo extensive manual curation.
They are thus generally assumed to be of high quality \citep{daneshjou_disparities_2022,daneshjou_skincon_2023}.
Nevertheless, as they contain a fairly limited number of samples, even few errors or inaccuracies can significantly impact evaluation results.
Moreover, besides public datasets, clinics frequently maintain private data collections that are instrumental for clinical research and for validating \gls*{ml} applications in local settings.
These collections can often not be shared or even leave premises due to privacy constraints.
The absence of truly standardized quality checks means that validation is left to the discretion of data owners, while the rest of the community (including reviewers) must rely on their diligence.

Recently, \gls*{dcai} initiatives have increased attention to data quality and produced modern tools for automatic analysis of image collections.
Prominent examples include \mbox{CleanLab}\footnote{\url{https://github.com/cleanlab/cleanvision}}, a toolbox that finds candidate label errors using confident learning \citep{northcutt_confident_2022} but can also identify several other issues in image collections ranging from out-of-distribution samples to extreme aspect ratios.
FastDup\footnote{\url{https://github.com/visual-layer/fastdup}} is a public library targeted at finding duplicates in large computer vision datasets, with additional features to identify sub-clusters, label errors, and other types of noise.

Very recently, SelfClean~\citep{groger_selfclean_2023} was introduced to investigate irrelevant samples, near duplicates, and label errors within the context of the dataset without relying on additional external data.
Its main difference from other existing approaches is that it reformulates dataset cleaning as a set of ranking problems instead of a classification problem.
These three issue types can lead to undesirable consequences when they affect evaluation sets.
Irrelevant samples are unsuitable to carry out valid tasks in the dataset context, and their presence introduces noise into the evaluation metrics.
Near duplicates are different views of the same object, which can create arbitrary re-weighting within the evaluation set or leak information when split over training and evaluation set.
Label errors, which encompass wrongly annotated samples, lead to inaccurate evaluations.
Throughout the remainder of this paper, we use the term ``noise'' to encompass all data quality issues, including these three main categories. 
Examples of each noise type in the context of dermatology are shown in Figure \ref{fig:Teaser}.

\paragraph{Contributions.}
This paper introduces a novel data-cleaning protocol that combines an existing method to find issues with an interpretable stopping criterion for efficient annotation (Section \ref{sub:Expert-IAA}).
The protocol is used to quantify the amount and type of noise in six popular dermatology evaluation datasets, which are then improved by rectifying issues when appropriate (Section \ref{sub:Identified-Issues}).
We then assess the influence of this data-cleaning effort on model performance estimates (Section \ref{sub:Influence-Cleaning}).
While focusing on specific datasets, we provide an effective, streamlined procedure to clean up image collections across the medical domain (Section \ref{sub:Quality-Rankings}), which is applicable even when data cannot leave private premises.
Furthermore, we investigate how non-expert annotators compare with domain experts in specific data quality confirmation tasks and find hints that near-duplicate detection does not require domain expertise (Section \ref{sub:Non-Expert-Comparison}).
Finally, we deliver the cleaned, more reliable benchmarks as ready-to-use file lists, improving the assessment of progress in the field.

\section{Related work}
The performance and ability of deep neural networks to make accurate predictions and generalize effectively relies heavily on the quality and quantity of available data \citep{rolnick_deep_2018}. 
However, real-world datasets frequently suffer from data quality issues.
In response to this challenge, a vast body of literature exists on mitigating the impact of noise during training \citep{karimi_deep_2020,natarajan_learning_2013,song2022learning}.
This can be achieved through various means, such as adapting architectures, utilizing specialized loss functions, or modifying training procedures.
However, these methods do not address the noise in the evaluation data, which affects the scores used to judge the model performance.
On the one hand, there are formidable efforts to curate evaluation datasets to very high-quality standards manually \citep{daneshjou_disparities_2022,irvin_chexpert_2019}.
This manual curation is challenging because humans are prone to errors, and it is already highly resource-intensive for mid-scale datasets \citep{pandey_modeling_2022}.
On the other hand, methods that automatically detect errors \citep{cheng_learning_2021,northcutt_confident_2022} and correct them either algorithmically or with reduced human effort \citep{northcutt_pervasive_2021} are applicable to evaluation as well as training sets.
Automatic detection approaches focus either on customized training processes to first learn with noisy supervision and then make decisions based on the output \citep{northcutt_confident_2022} or on directly using features learned without labels for decision making \citep{groger_selfclean_2023}.

Recently, several works investigated and produced cleaned versions of existing benchmarks for general computer vision \citep{northcutt_pervasive_2021}, intent detection \citep{ying_label_2022}, and automated electrocardiogram interpretation \citep{doggart_automated_2022}.
These efforts mostly use automated noise detection frameworks such as confident learning \citep{northcutt_confident_2022} followed by human confirmation.
In the domain of dermatology, \citet{cassidy_analysis_2022} analyzed the prevalence of near duplicates in the \gls*{isic} datasets and provided practical recommendations on their usage for reliable evaluation.
However, most of these endeavors primarily focus on a single noise type, while different kinds of data quality issues may be present at once in a dataset, with significant impact for \acrlong*{ml}.

\section{Methodology}

\subsection{Problem statement}
Data cleaning refers to the process of identifying, correcting, and removing errors, inconsistencies, and inaccuracies from raw data collections to enhance their reliability and usability \citep{li_cleanml_2021}. 
It involves a number of tasks aimed at improving data quality, such as addressing outliers, duplicate records, and many other modality-specific issues. 
The ultimate goal of data cleaning is to prepare datasets for analysis and decision-making, ensuring that the insights drawn from the data are accurate and trustworthy.
In this paper, we focus on three data quality issues, which we empirically found to be frequent in medical imaging benchmark datasets and, at the same time, difficult to detect.
These are irrelevant samples, near duplicates, and label errors.

The primary emphasis of this work lies in the detection and treatment of these quality issues rather than in the comparison of existing detection strategies.
Thus, we rely on a single cleaning tool in combination with expert confirmation.

\subsection{Ranking candidate data quality issues}
We leverage SelfClean \citep{groger_selfclean_2023} to identify potential data quality issues in dermatology benchmark datasets.
SelfClean is a holistic data-cleaning framework based on recent advances in \gls*{ssl} \citep{ozbulak2023know} that targets the detection of irrelevant samples, near duplicates, and label errors.
It achieves remarkable performance on small- to medium-size datasets without leveraging labels or additional data sources.

SelfClean first uses self-supervised pre-training on a noisy dataset to obtain an encoder that maps samples onto a dataset-specific latent space.
The method can be applied in combination with any \gls*{ssl} approach, but the original work compared \mbox{\gls*{simclr}}~\citep{chen_simple_2020} and \gls*{dino}~\citep{caron_emerging_2021} and found the latter to have better performance.
We refer to \citet{groger_selfclean_2023} for ablations on the influence of different pre-training strategies.
The dataset-specific latent space is then exploited to find data quality issues using simple distance-based criteria.
Specifically, single-linkage agglomerative clustering is used to find irrelevant samples, pairwise distances are computed to identify near duplicates, and the intra-/extra- class distance ratio is used to detect label errors.
For each noise type, SelfClean yields a sorted list of all samples (or pairs thereof), where items appearing earlier in the sequence are more likely to manifest data quality issues.

\subsection{Confirming data quality issues} \label{sub:DQI}
We present the rankings obtained with SelfClean for each dataset and noise type to human annotators for confirmation.
Note that this manual cleaning, which involves human feedback, is recommended in \citet{groger_selfclean_2023} for evaluation sets,
where it is crucial to control potential score bias that comes from using similar methods for cleaning and performing the final classification task.
We rely on annotations by three practicing domain experts (referred to as E1, E2, and E3), of which E3 is a board-certified dermatologist, and leverage the same verification tool developed in \citet{groger_selfclean_2023}.
Experts select a dataset and data quality issue type, then answer the binary questions reported in Appendix \ref{app:Human-Confirmation}.

Comprehensive verification of potential issues for all samples $N$ in a dataset presents a challenge due to the high effort required and may be ultimately detrimental due to natural limitations in the attention span of annotators.
This is exacerbated in the case of near duplicates, where a binary choice should be taken for each of the $N(N-1)/2$ possible pairs of images.
Thus, there is a trade-off between the resources allocated to annotation and the number of data quality issues that remain unverified.
To solve this issue, we propose a conservative and intuitive stopping criterion to terminate the annotation process before the dataset is exhaustively covered.
Specifically, we proceed along the SelfClean ranking and stop the annotation process after receiving $n_{\text{clean}}$ consecutive negative responses.
We set $n_{\text{clean}}$ by requesting that the probability of observing the sequence of negative annotations as a result of chance be lower than $p_\text{chance}$,
where the probability for each sample to be a data quality issue is $p_+$ (or less).
Thus the probability of observing a sequence of $n_\text{clean}$ clean samples in a row by chance is $p_\text{seq}=(1-p_+)^{n_\text{clean}}$.
The observation of such a sequence is therefore significant with 95\% confidence when $p_\text{seq}\le 5\%=p_\text{chance}$,
corresponding to $n_\text{clean}\ge\ln p_\text{chance}/\ln(1-p_+)$.
For simplicity, $p_+$ is set to a fixed typical value and is not estimated from the sequence.
Its dependence on the ranking is ignored, and no correction for the observation of multiple sequences is applied.
This yields $n_{\text{clean}} = \lfloor \ln(p_\text{chance})/\ln(1 - p_+) \rfloor$, where we set $p_\text{chance} = p_+ = 0.05$, resulting in $n_{\text{clean}} = 58$.

\subsection{Cleaning data quality issues} \label{sub:Cleaning}
After the confirmation process, we conservatively require unanimous expert agreement to identify an issue.
Specifically, a sample is considered noise only if all experts flag it as such. 
We then produce cleaned benchmark datasets by discarding confirmed irrelevant samples and randomly removing a sample for each confirmed pair of near duplicates.
We note that this treatment of near-duplicates is appropriate only when there are at most two images of the same object and different views are approximately equivalent, which we found to be the case for the considered datasets.
The updated file lists include the file names of valid images from the original dataset, excluding irrelevant and near-duplicate samples.
Furthermore, we estimate the percentage of label errors from expert confirmations.
Label errors are not corrected as this would unfairly favor models similar to the SelfClean encoder.
In fact, since only the first part of the sequence is proposed for verification, samples which would be incorrectly classified by nearest neighbors on SelfClean representations are more likely to be removed from the dataset, which would positively bias the score.
Instead, the estimated prevalence of label errors is reported.
When models achieve an error rate comparable to this estimate or lower, there is an indication that performance may be limited by label quality and further optimization could lead to overfitting.
Moreover, some of the datasets obtain their labels as a result of pathological confirmation.
Thus, correction on the basis of visual appearance would be inappropriate.

\begin{figure*}[t]
\floatconts
  {fig:IAA-Overview}
  {\caption{
      Inter-annotator agreement as Krippendorff's alpha among all expert annotators (left) and Cohen's kappa for all expert annotator pairs (right).
      Markers identify the six selected evaluation datasets,
      error bars are 95\% confidence intervals obtained by bootstrapping annotated samples,
      and the background color indicates the degree of agreement \citep{regier_dsm-5_2013}.
  }}
  {\includegraphics[width=1.0\linewidth]{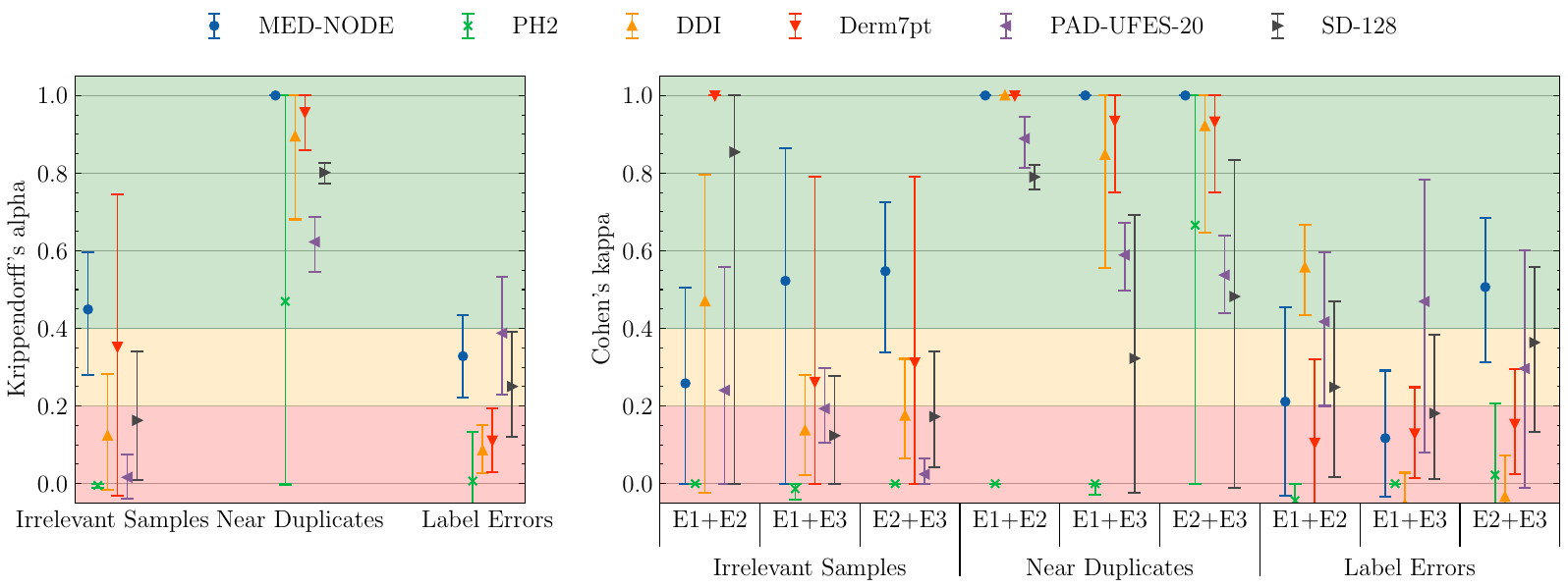}}
\end{figure*}

\subsection{Evaluation benchmark dataset selection}
Six popular small-scale dermatology datasets are selected for cleaning from a list of public sources for skin image analysis research by the \gls*{isic} community\footnote{\url{https://workshop2023.isic-archive.com}}.
These can be considered evaluation datasets as their size makes them poorly suited for training, and they usually cover a specific use case of interest.
Public datasets without licenses fall under the public domain mark.

\begin{description}
    \item[MED-NODE~\citep{giotis2015a}] features 170 clinical images for skin cancer detection by the University Medical Center Groningen, Netherlands (CC BY 4.0). The images are categorized into melanoma and naevus.
    
    \item[PH2~\citep{mendonca_ph2_2013}] features 200 dermoscopy images for melanocytic lesion classification by the Hospital Pedro Hispano in Matosinhos, Portugal. The images are categorized into common nevi, atypical nevi, and melanomas.
    
    \item[DDI~\citep{daneshjou_disparities_2022}] features 656 clinical images for skin cancer detection and rare disease classification by the Stanford Clinics (Stanford's University dataset research agreement). The images are categorized into benign or malignant lesions and 78 diseases.
    
    \item[Derm7pt~\citep{Kawahara2018-7pt}] features 2,022 dermoscopy and clinical images for skin condition diagnosis (CC BY-NC-SA 4.0). The images are categorized into 16 diagnoses.
    
    \item[PAD-UFES-20~\citep{pacheco_pad-ufes-20_2020}] features 2,298 clinical images for skin condition diagnosis (CC BY 4.0). The images are categorized into six disease diagnoses.
    
    \item[SD-128~\citep{leibe_benchmark_2016}] features 5,619 clinical images for skin condition diagnosis collected from DermQuest. The images are categorized into 123 diseases.
\end{description}

\begin{table*}[t]
\centering
\floatconts
  {tab:Identified-Issues}
  {\caption{
    Expert-confirmed data quality issues found in six dermatology benchmarks,
    for all three noise types and two aggregation strategies.
  }}
  {\begin{tabular}{l  c  rrr  rrr}
  \toprule
        \multicolumn{2}{c}{}
            & \multicolumn{3}{c}{\textbf{Majority Voting}}
            & \multicolumn{3}{c}{\textbf{Unanimous Agreement}} \\
        \cmidrule(lr){3-5}
        \cmidrule(lr){6-8}
        &
            & \multicolumn{1}{c}{Irrelevant}
            & \multicolumn{1}{c}{Near}
            & \multicolumn{1}{c}{Label}
            
            & \multicolumn{1}{c}{Irrelevant}
            & \multicolumn{1}{c}{Near}
            & \multicolumn{1}{c}{Label} \\[-1pt]
        \textbf{Dataset}
        & \textbf{Size}
            & \multicolumn{1}{c}{Samples}
            & \multicolumn{1}{c}{Duplicates}
            & \multicolumn{1}{c}{Errors}
            
            & \multicolumn{1}{c}{Samples}
            & \multicolumn{1}{c}{Duplicates}
            & \multicolumn{1}{c}{Errors}\\
    \midrule
        MED-NODE & 170 
            & 8 (4.7\%)
            & 1 (0.6\%)
            & 15 (8.8\%)
            & 3 (1.8\%)
            & 1 (0.6\%)
            & 2 (1.2\%) \\
            
        PH2 & 200 
            & 0 (0.0\%)
            & 0 (0.0\%)
            & 0 (0.0\%)
            & 0 (0.0\%)
            & 0 (0.0\%)
            & 0 (0.0\%) \\
            
        DDI & 656 
            & 8 (1.2\%)
            & 6 (0.9\%)
            & 43 (6.6\%)
            & 3 (0.5\%)
            & 6 (0.9\%)
            & 8 (1.2\%) \\
            
        Derm7pt & 2,022 
            & 1 (0.1\%)
            & 9 (0.5\%)
            & 15 (0.7\%)
            & 1 (0.1\%)
            & 9 (0.5\%)
            & 2 (0.1\%) \\
            
        PAD-UFES-20 & 2,298 
            & 12 (0.5\%)
            & 66 (2.9\%)
            & 8 (0.3\%)
            & 2 (0.1\%)
            & 56 (2.4\%)
            & 3 (0.1\%) \\
            
        SD-128 & 5,619 
            & 4 (0.1\%)
            & 167 (3.0\%)
            & 12 (0.2\%)
            & 3 (0.1\%)
            & 156 (2.8\%)
            & 4 (0.1\%) \\
    \bottomrule
    \end{tabular}}
\end{table*}

\subsection{Implementation details}
We train a randomly initialized \acrlong*{vit} tiny encoder with a patch size of $16 \times 16$ using \gls*{dino} on each considered dataset separately.
The latent representation is given by the class token output from the encoder's last layer, which has dimension 192.
\Gls*{ssl} pre-training was performed for 500 epochs.
Similar to \citet{groger_selfclean_2023}, we use stronger data augmentation than the original \gls*{dino} strategy \citep{caron_emerging_2021} to make the self-supervised task sufficiently complex.
All images are resized to $224 \times 224$ pixels and normalized using mean and standard deviation of ImageNet \citep{deng_imagenet_2009}.
Our implementation is based on PyTorch~\citep{paszke_pytorch_2019} and the official SelfClean repository \citep{groger_selfclean_2023}.
All experiments are performed on an \mbox{Nvidia DGX-1} server.

\section{Results}

\subsection{Expert inter-annotator agreement} \label{sub:Expert-IAA}
Figure \ref{fig:IAA-Overview} presents an analysis of the inter-annotator agreement of the experts for each dataset and noise type.
In the left panel, the overall agreement for all three expert annotators is measured with Krippendorff's alpha, which naturally allows for different sets of labeled samples.
In the right panel, each pair of expert annotators is compared using Cohen's kappa, computed on all samples labeled by both.
Error bars represent 95\% confidence intervals obtained with 1,000 bootstrap repetitions on the annotated samples.
Background colors indicate the degree of agreement according to the ranges defined in~\citet{regier_dsm-5_2013}, i.e. good ($>$0.4), questionable (0.2--0.4), and unacceptable ($<$0.20).

We observe that the agreement for the considered noise types is vastly different.
Expert annotators largely agree on near duplicates across most datasets.
On the other hand, agreement is lower for label errors and irrelevant samples.
This behavior is expected since these two tasks are not direct image comparisons and involve applying dermatological criteria outside of a clinical context.
As a consequence, complexity and subjectivity are higher, especially for rare cases \citep{daneshjou_disparities_2022}.
Furthermore, our findings align with those of \citet{daneshjou_disparities_2022}, which showed that even board-certified dermatologists had difficulties identifying whether lesions were benign or malignant purely visually, underscoring the importance of labeling with pathology results.
The agreement is significantly lower for datasets containing dermoscopy images (PH2 and Derm7pt) than for those with clinical images, an interesting finding as dermoscopy is generally perceived to improve diagnostic accuracy \citep{kittler_diagnostic_2002}.
Possible root causes are magnification and standardization, which deprive expert annotators of additional informative visual features that are especially relevant for harder cases.

\subsection{Identified data quality issues} \label{sub:Identified-Issues}
Table \ref{tab:Identified-Issues} quantifies the data quality issues found in each dataset for each noise type after aggregating annotations using majority voting or unanimous agreement.
Since we require the agreement of multiple experts, we expect the subjectivity of annotations to be reduced for confirmed problems.
In other words, low agreement leads to fewer issues compared to those reported by individual annotators.
This is apparent for the two dermoscopy datasets, PH2 and Derm7pt.
Even with the strictest criterion of unanimous agreement, we observe that the number of near duplicates found increases with dataset size up to 2--3\% for these already highly curated datasets, indicating that avoiding this type of error becomes more difficult.
We also identify up to 2\% of irrelevant samples and estimate label errors to be approximately 1\% for DDI and MED-NODE, potentially related to edge cases.
PH2 is the only dataset where we do not find any data quality issues.
Detected issues can be visually inspected in Appendix \ref{app:Found-Issues}.

Figure \ref{fig:Expert-Verified} in Appendix \ref{app:Detailed-Cleaning} shows the number of items inspected by the experts for each dataset.
For most of the noise types, datasets, and experts, the stopping criterion terminated the annotation process (see Section \ref{sub:DQI}).
On average, this speeds up the annotation process by a factor of 14 for irrelevant samples, 32,520 for near duplicates, and 12 for label errors compared to exhaustive annotation.
Table \ref{tab:Speed-Up-Annotation} in Appendix \ref{app:Detailed-Cleaning} gives a breakdown of these figures for each dataset and noise type.
The remarkable speed-up factor for near duplicates illustrates how resource-intensive such detection is and how it can be significantly reduced with the proposed data-cleaning protocol.
To showcase the robustness of the approach, we investigate the dependence of the number of detected samples on the two parameters $p_+$ and $p_\text{chance}$ of the stopping criterion in Appendix \ref{app:Ablation-Stopping}.
Results indicate that the proposed criterion is largely insensitive to the choice of parameters for most issue types and datasets.

\subsection{Influence of cleaning data quality issues} \label{sub:Influence-Cleaning}
Table \ref{tab:Influence-DataQuality} showcases the impact of data cleaning on the performance of two prominent open-source binary classification models designed for skin cancer detection: DeepDerm \citep{esteva_dermatologist-level_2017} and HAM10000 \citep{tschandl_ham10000_2018}. 
These models are pre-trained on external data, and we evaluate them on all considered datasets before and after cleaning.
We then report the performance difference in terms of \gls*{auroc}, \gls*{ap}, and \gls*{auprg}.
Note that the cleaning process only involves the removal of irrelevant samples and near duplicates (see Section \ref{sub:Cleaning}).
Despite the limited number of issues found within the original datasets, their influence on model performance estimates remains evident.
The performance difference observed for the DDI dataset in DeepDerm is particularly noteworthy, as there is a significant performance reduction of $-1.0\%$, $-1.4\%$, and $-3.6\%$ in \gls*{auroc}, \gls*{ap}, and \gls*{auprg}, respectively.
For \mbox{MED-NODE}, both models observe a noteworthy reduction in \gls*{auprg}. 
Additionally for SD-128 there is a mild but significant difference for DeepDerm in \gls*{auroc} and \gls*{ap} and in \mbox{\gls*{auroc}} for HAM10000.
Overall, these findings underscore the significance of data cleaning and its influence on the assessment and selection of models.

\begin{table*}[t]
\centering
\floatconts
  {tab:Influence-DataQuality}
  {\caption{
    Difference in performance after cleaning evaluation datasets for two open-source binary classification models for skin cancer detection trained on external data.
    Performance is compared in terms of \gls*{auroc}, \gls*{ap}, and \gls*{auprg}.
    PH2 is not evaluated, as it was found to contain no issues.
    We report the median and the 95\% confidence interval in brackets.
    Additionally $^*$ represents significance when zero is strictly outside the range, and $^\circ$ denotes borderline cases when a boundary is exactly zero.
  }}
  {
  \resizebox{\linewidth}{!}{%
  \begin{tabular}{l  lll  lll}
  \toprule
        \multicolumn{1}{c}{}
            & \multicolumn{3}{c}{\textbf{DeepDerm}}
            & \multicolumn{3}{c}{\textbf{HAM10000}} \\
        \cmidrule(lr){2-4}
        \cmidrule(lr){5-7}
            & \multicolumn{1}{c}{Difference}
            & \multicolumn{1}{c}{Difference}
            & \multicolumn{1}{c}{Difference}
            
            & \multicolumn{1}{c}{Difference}
            & \multicolumn{1}{c}{Difference}
            & \multicolumn{1}{c}{Difference} \\[-1pt]
        \textbf{Dataset}
            & \multicolumn{1}{c}{AUROC (\%)}
            & \multicolumn{1}{c}{AP (\%)}
            & \multicolumn{1}{c}{AUPRG (\%)}
            
            & \multicolumn{1}{c}{AUROC (\%)}
            & \multicolumn{1}{c}{AP (\%)}
            & \multicolumn{1}{c}{AUPRG (\%)}\\
    \midrule
        MED-NODE 
            & \footnotesize $-0.3 \ [-0.7, +0.1]$ 
            & \footnotesize $-0.1 \ [-0.9, +0.4]$ 
            & \footnotesize $-0.9 \ [-2.5, \phantom{+}0.0]^\circ$
            
            & \footnotesize $-0.2 \ [-0.8, +0.2]$ 
            & \footnotesize $-0.0 \ [-0.9, +0.6]$ 
            & \footnotesize $-1.1 \ [-3.1, -0.0]^*$ \\
            
        DDI 
            & \footnotesize $-1.0 \ [-1.9, -0.2]^*$ 
            & \footnotesize $-1.4 \ [-3.4, -0.2]^*$ 
            & \footnotesize $-3.6 \ [-7.9, -0.7]^*$
            
            & \footnotesize $+0.1 \ [-0.9, +1.0]$ 
            & \footnotesize $-0.6 \ [-2.6, +0.3]$ 
            & \footnotesize $+0.7 \ [-2.0, +3.9]$ \\
            
        Derm7pt
            & \footnotesize $+0.1 \ [-0.1, +0.4]$ 
            & \footnotesize $-0.1 \ [-0.4, \phantom{+}0.0]^\circ$ 
            & \footnotesize $+0.4 \ [-0.1, +1.0]$
            
            & \footnotesize $+0.0 \ [-0.2, +0.3]$ 
            & \footnotesize $-0.1 \ [-0.5, +0.2]$ 
            & \footnotesize $+0.4 \ [-0.3, +1.1]$ \\
            
        PAD-UFES-20
            & \footnotesize $+0.1 \ [-0.4, +0.7]$ 
            & \footnotesize $+0.1 \ [-0.1, +0.5]$ 
            & \footnotesize $-0.3 \ [-1.2, +0.6]$
            
            & \footnotesize $+0.1 \ [-0.4, +0.7]$ 
            & \footnotesize $+0.1 \ [-0.2, +0.5]$ 
            & \footnotesize $+0.1 \ [-0.7, +0.8]$ \\
            
        SD-128
            & \footnotesize $-0.1 \ [-0.3, -0.0]^*$ 
            & \footnotesize $+0.3 \ [+0.0, +0.5]^*$ 
            & \footnotesize $-0.1 \ [-0.4, +0.3]$
            
            & \footnotesize $-0.2 \ [-0.6, -0.0]^*$ 
            & \footnotesize $+0.1 \ [-0.0, +0.1]$ 
            & \footnotesize $-0.9 \ [-2.1, +0.2]$ \\
    \bottomrule
    \end{tabular}}
    }
\end{table*}

\subsection{Quality of candidate rankings} \label{sub:Quality-Rankings}
Table \ref{tab:Quality-Rankings} in Appendix \ref{app:Detailed-Cleaning} compares the candidate rankings produced by SelfClean and the unanimous agreement of expert annotators for samples annotated by all three.
Performance is measured in terms of \gls*{auroc}, \gls*{ap}, and \gls*{auprg} \citep{flach_precision-recall-gain_2015}.
Results for irrelevant samples indicate that the ranking is well-aligned with expert annotations except for SD-128, where scores indicate no advantage.
The agreement of near duplicates suggested by SelfClean with expert annotations is the highest among the three noise types, as indicated by all scores always well above the baselines.
Label errors seem the most difficult for the chosen data-cleaning support strategy, with scores at baseline level for Derm7pt and marginal gain for SD-128.
The poor performance that we observe in a limited number of cases could be due to the intrinsic difficulty of the task, but also to the small support of problematic examples in already highly curated evaluation datasets.
Overall there is good correspondence in our experiments between the SelfClean rankings and expert-confirmed data quality issues.
However, we also see that relying solely on rankings can lead to suboptimal cleaning, reinforcing the necessity for human confirmation.

\subsection{Non-expert confirmation} \label{sub:Non-Expert-Comparison}
The proposed protocol has been shown to speed up the cleaning process significantly. 
However medical expert annotations are very expensive to obtain.
Thus, here we want to test if in-depth medical knowledge is needed for all three confirmation tasks, as our protocol would then yield a speed-up and simultaneously reduce costs.
Here we recruited three laypersons who repeated the annotation tasks for MED-NODE, PH2, and \mbox{SD-128}.
Results are reported in Figure \ref{fig:IAA-Expert-Lay}.
The upper panels show the pair-wise Cohen kappas for each expert-expert and expert-lay pair.
The lower panels show the paired differences between the agreement of every expert and layperson with the same reference expert.
We test for statistical significance of two-expert pairs having a better agreement than expert-lay pairs with a one-sided paired permutation test, which solely relies on the sign of the differences.
The results indicate no sizeable difference between experts and laypersons for irrelevant sample detection except SD-128, which might be a consequence of poor expert agreement.
Although experts agree better among each other for near duplicates on MED-NODE and PH2, this is not particularly noteworthy as the support is only a few samples.
On the other hand, for SD-128, experts and laypersons can identify near-duplicate samples in the hundreds without a significant performance difference.
These results suggest that non-experts may be effective at annotating near duplicates, although further investigation with more datasets and subjects is needed. 
We note however that similar observations were made for estimating Fitzpatrick skin types with crowd workers \citep{groh2022towards}.
In the case of label errors, experts have an edge over laypersons as expected.

\begin{figure*}[t]
\floatconts
  {fig:IAA-Expert-Lay}
  {\caption{
        The upper panels show the pair-wise annotator agreement using Cohen's kappa of each expert-expert (blue) and expert-lay person pair (orange).
        The lower panels show the difference in Cohen's kappa between every pair of an expert and a layperson with respect to the same reference expert.
  }}
  {%
    \includegraphics[width=\textwidth]{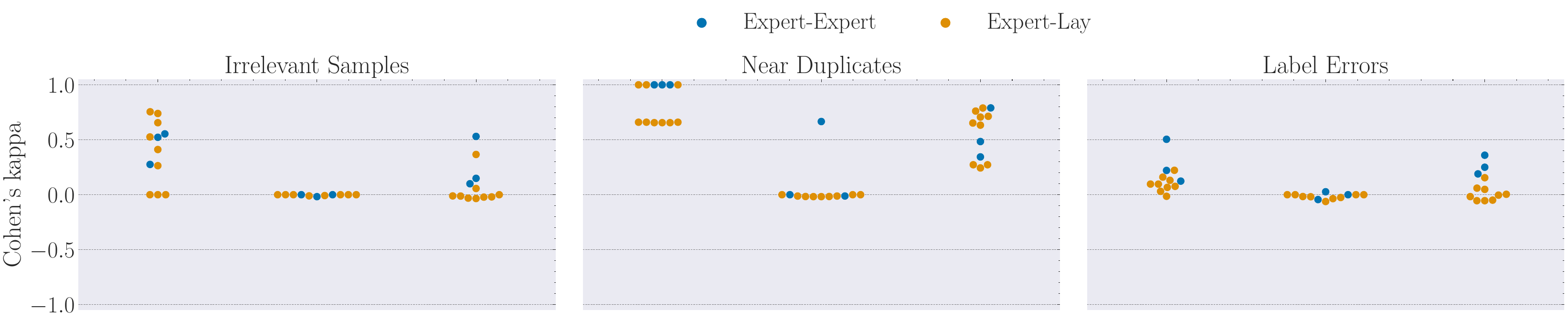}
    \includegraphics[width=\textwidth]{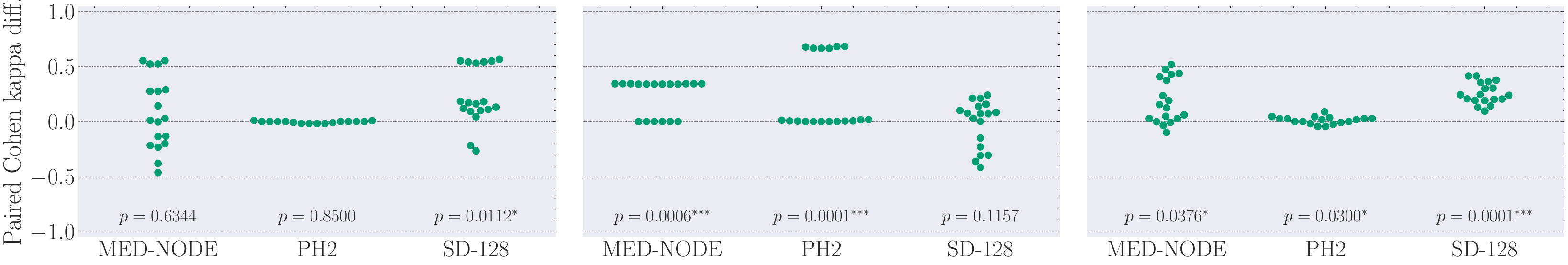}
  }
\end{figure*}

\section{Conclusion}
This paper outlined the application of a novel efficient data-cleaning protocol for irrelevant samples, near duplicates, and label errors,
which combines domain-expert confirmation of algorithmically ranked candidates with an intuitive stopping criterion.
The procedure was applied to six popular public evaluation datasets for digital dermatology to produce more reliable performance measurements by removing issues universally recognized by three experts.
We removed up to 1.8\% of irrelevant samples, up to 2.8\% of near duplicates, and estimated label error prevalence in all considered datasets with a significant speed-up compared to exhaustive annotation.
We found performance differences up to $-1.0\%$ in AUROC, $-1.4\%$ in AP, and $-3.6\%$ in AUPRG when evaluating models on cleaned versions of these already highly curated datasets.
The outlined data-cleaning procedure demonstrated remarkable effectiveness and robustness, and may be used to improve public and private datasets including medical atlases independently of the domain.
Finally, we found hints that near-duplicate confirmation does not require domain expertise, which can significantly lower the cost of curation.
Revised file lists are available at \url{https://github.com/Digital-Dermatology/SelfClean-Revised-Benchmarks}. 

By eliminating label errors, irrelevant samples, and near duplicates, the improved datasets obtain more realistic performance estimates for clinical decision support systems.
This not only streamlines the diagnostic process but also minimizes the risk of misdiagnosis, thereby improving patient outcomes. 
The impact of the outlined procedure is expected to be larger on uncurated datasets as they usually contain more inaccuracies, and no other approach exists than extensive manual curation.

In future work, we aim to clean larger datasets including development and evaluation splits to make \gls*{ml} more reliable in digital dermatology.

\acks{
    We want to thank Xu Cao, who supported us as a mentor during the final stages of paper writing, and Wenqian Ye for reviewing our paper.
    Additionally, we want to thank all the anonymous reviewers for their insightful and thoughtful comments.
}

\bibliography{groeger23}

\clearpage
\appendix

\section{Limitations}
One limitation of the current data-cleaning protocol is that the annotation process is influenced by the rankings produced by SelfClean.
Recent works have found that human annotation quality depends on the order of the instances shown to the annotators \citep{pandey_modeling_2022}.
In its present form, the procedure used in this paper still features a complicated interplay between the proposed ranking, deviations from the ranking, and the quality of annotations.

The proposed data-cleaning protocol obtains candidate rankings using SelfClean, which relies on \gls*{ssl} and thus does not inherit annotation biases.
These rankings are then combined with the in-depth domain knowledge of multiple human experts to find data quality issues in existing benchmark datasets.
Although we leverage multiple experts, the confirmed data quality issues may still be affected by human subjectivity and biases.
However, we mitigate this as much as possible by requiring unanimous agreement.

Label errors may be difficult to confirm, as atypical presentations of disease, such as amelanotic melanomas or pigmented non-melanoma skin cancers, may be flagged as a label error when the label is in fact correct.
Prior work has shown that even board-certified dermatologists have difficulty reaching a consensus on difficult cases where pathological confirmation provides a gold-standard label \citep{daneshjou_disparities_2022,daneshjou_skincon_2023}.
Labels obtained through procedures such as biopsies or follow-up consultations should \emph{not} be corrected by human image assessment as they represent a different source of truth.
This is one of the reasons why the proposed protocol only estimates label error prevalence, leaving further investigation to the dataset curators.

\section{Illustration of detected data quality issues}\label{app:Found-Issues}
This section illustrates the data quality issues found in the considered datasets by unanimous expert agreement.
The figures \ref{fig:MedNode-Issues}, \ref{fig:DDI-Issues}, \ref{fig:derm7pt-Issues}, \ref{fig:PAD-UFES-20-Issues}, and \ref{fig:SD-128-Issues} show up to seven issues for each dataset, with the exception of PH2, as it was found not to contain any issues.

\begin{figure*}[htbp]
\floatconts
  {fig:MedNode-Issues}
  {\caption{
    Visualization of the data quality issues found in MED-NODE by unanimous expert agreement, i.e. 3 irrelevant samples, 1 near duplicate, and 2 label errors.
    Figure \ref{fig:MedNode-Irrelevants} shows the irrelevant samples, \ref{fig:MedNode-Dups} the near duplicates, and \ref{fig:MedNode-Labels} the label errors.
  }}
  {%
    \subfigure[Irrelevant Samples]{\label{fig:MedNode-Irrelevants}%
      \includegraphics[width=1.0\linewidth]{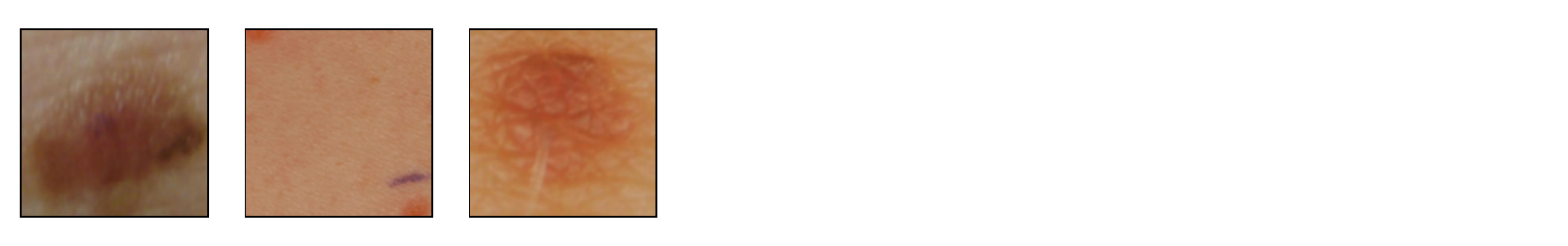}}%
      
    \subfigure[Near Duplicates]{\label{fig:MedNode-Dups}%
      \includegraphics[width=1.0\linewidth]{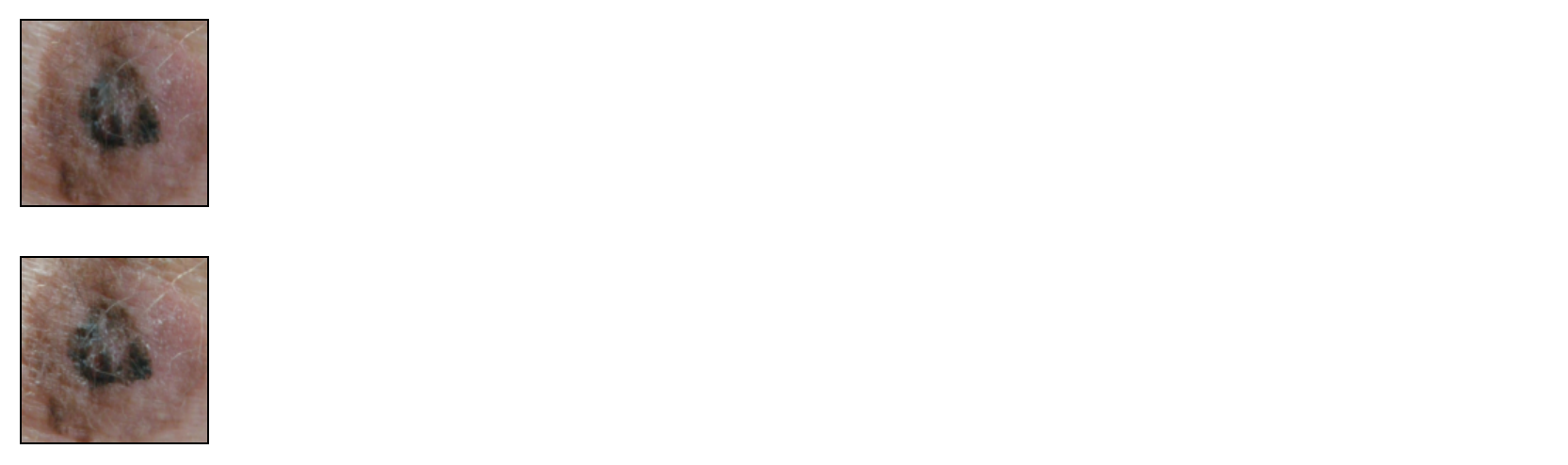}}

    \subfigure[Label Errors]{\label{fig:MedNode-Labels}%
      \includegraphics[width=1.0\linewidth]{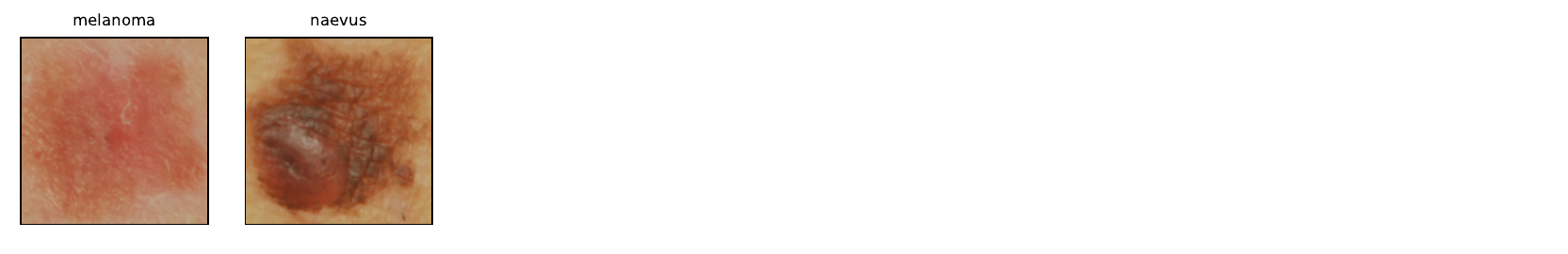}}
  }
\end{figure*}

\begin{figure*}[htbp]
\floatconts
  {fig:DDI-Issues}
  {\caption{
    Visualization of the data quality issues found in DDI by unanimous expert agreement, i.e. 3 irrelevant samples, 6 near duplicates, and 8 label errors.
    Figure \ref{fig:DDI-Irrelevants} shows the irrelevant samples, \ref{fig:DDI-Dups} the near duplicates, and \ref{fig:DDI-Labels} the first 7 label errors.
  }}
  {%
    \subfigure[Irrelevant Samples]{\label{fig:DDI-Irrelevants}%
      \includegraphics[width=1.0\linewidth]{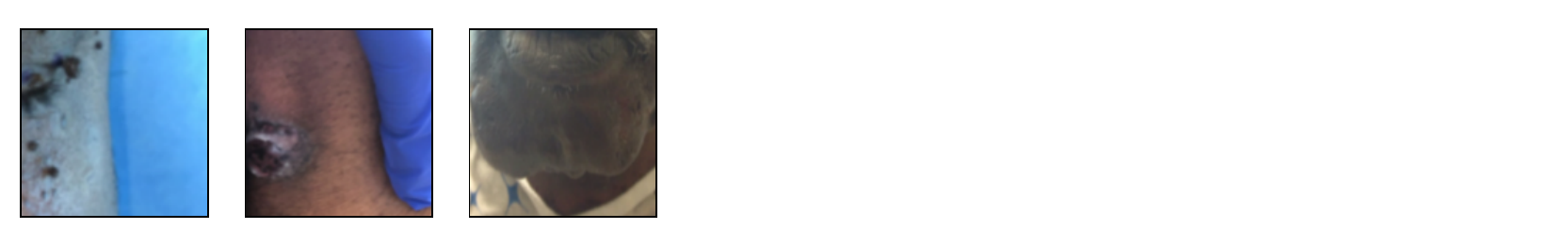}}%
      
    \subfigure[Near Duplicates]{\label{fig:DDI-Dups}%
      \includegraphics[width=1.0\linewidth]{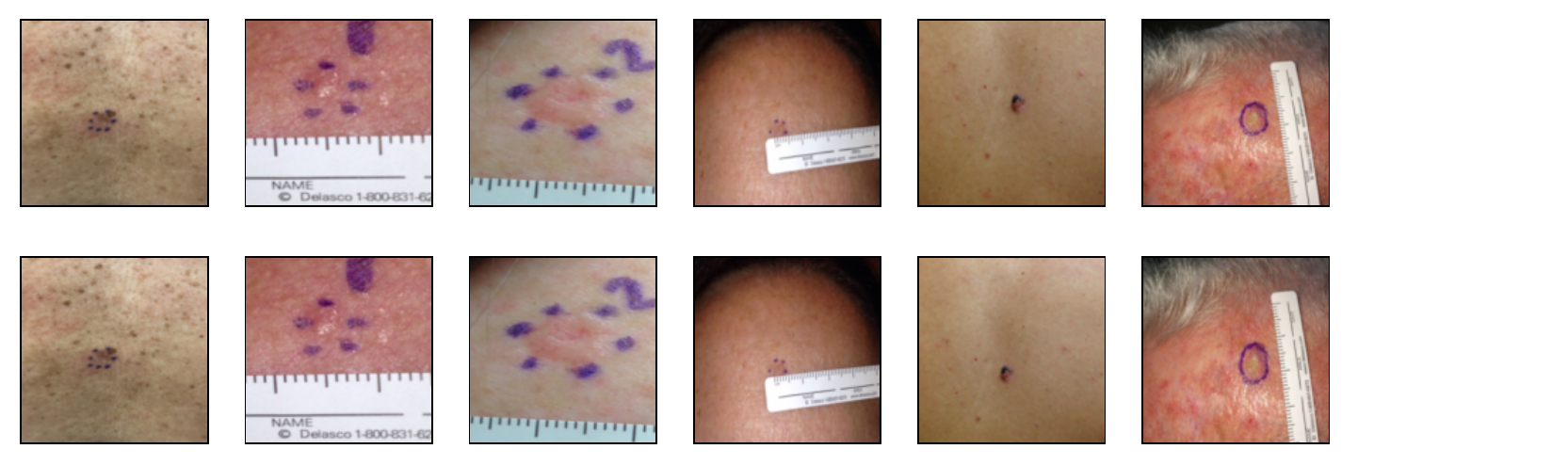}}

    \subfigure[Label Errors]{\label{fig:DDI-Labels}%
      \includegraphics[width=1.0\linewidth]{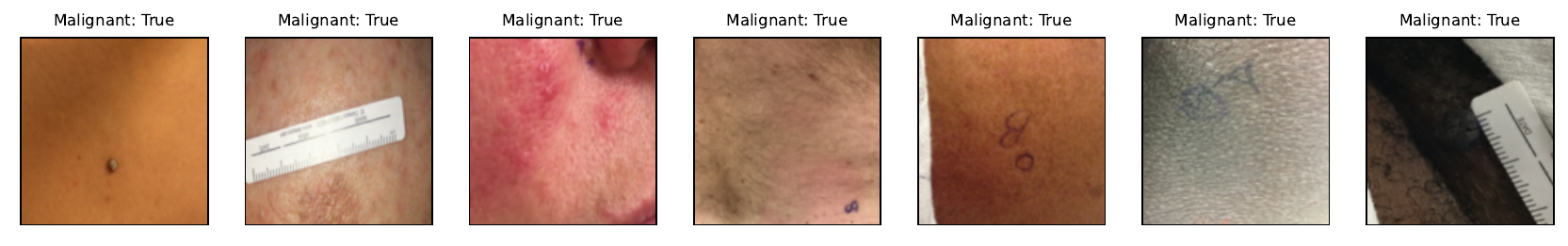}}
  }
\end{figure*}

\begin{figure*}[htbp]
\floatconts
  {fig:derm7pt-Issues}
  {\caption{
    Visualization of the data quality issues found in Derm7pt by unanimous expert agreement, i.e. 1 irrelevant sample, 9 near duplicates, and 2 label errors.
    Figure \ref{fig:derm7pt-Irrelevants} shows the irrelevant samples, \ref{fig:derm7pt-Dups} the first 7 near duplicates, and \ref{fig:derm7pt-Labels} the label errors.
  }}
  {%
    \subfigure[Irrelevant Samples]{\label{fig:derm7pt-Irrelevants}%
      \includegraphics[width=1.0\linewidth]{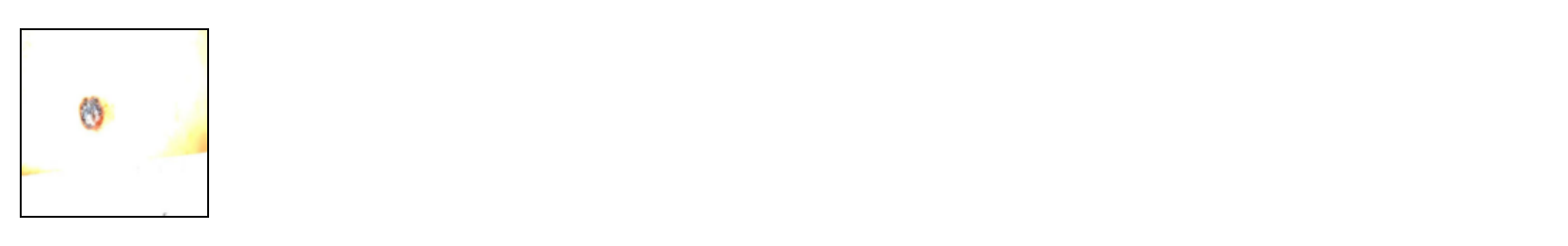}}%
      
    \subfigure[Near Duplicates]{\label{fig:derm7pt-Dups}%
      \includegraphics[width=1.0\linewidth]{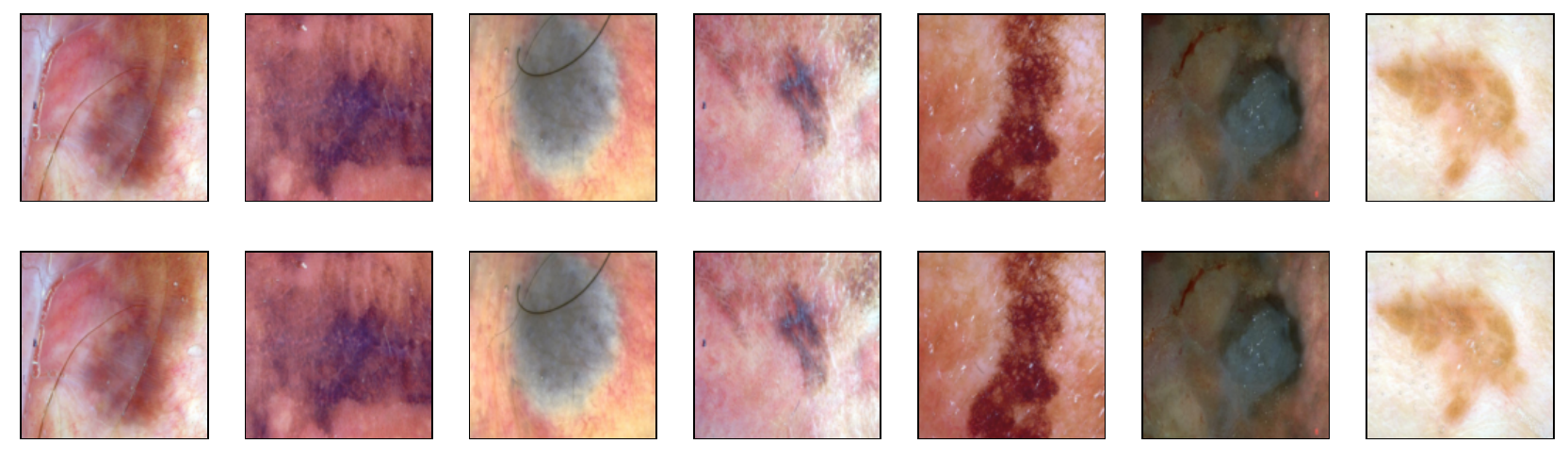}}

    \subfigure[Label Errors]{\label{fig:derm7pt-Labels}%
      \includegraphics[width=1.0\linewidth]{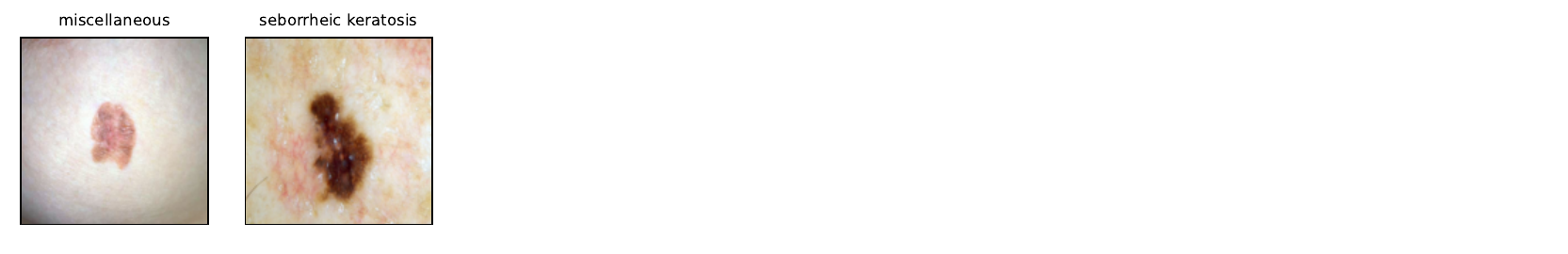}}
  }
\end{figure*}

\begin{figure*}[htbp]
\floatconts
  {fig:PAD-UFES-20-Issues}
  {\caption{
    Visualization of the data quality issues found in PAD-UFES-20 by unanimous expert agreement, i.e. 2 irrelevant samples, 56 near duplicates, and 3 label errors.
    Figure \ref{fig:PAD-UFES-20-Irrelevants} shows the irrelevant samples, \ref{fig:PAD-UFES-20-Dups} the first 7 near duplicates, and \ref{fig:PAD-UFES-20-Labels} the label errors.
  }}
  {%
    \subfigure[Irrelevant Samples]{\label{fig:PAD-UFES-20-Irrelevants}%
      \includegraphics[width=1.0\linewidth]{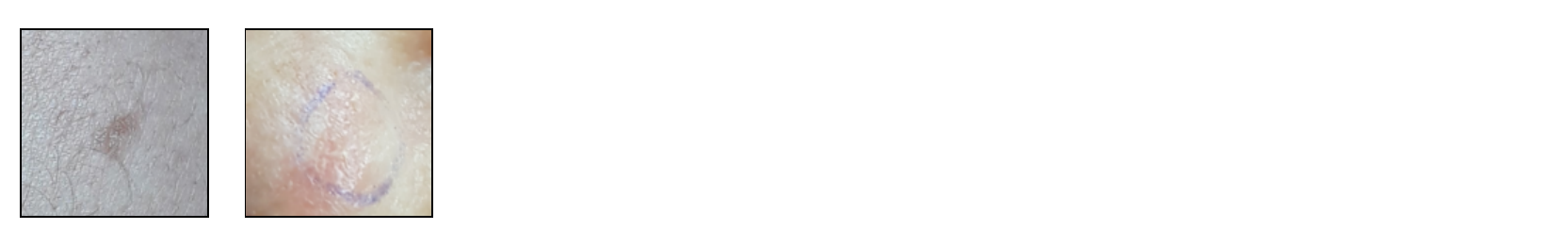}}%
      
    \subfigure[Near Duplicates]{\label{fig:PAD-UFES-20-Dups}%
      \includegraphics[width=1.0\linewidth]{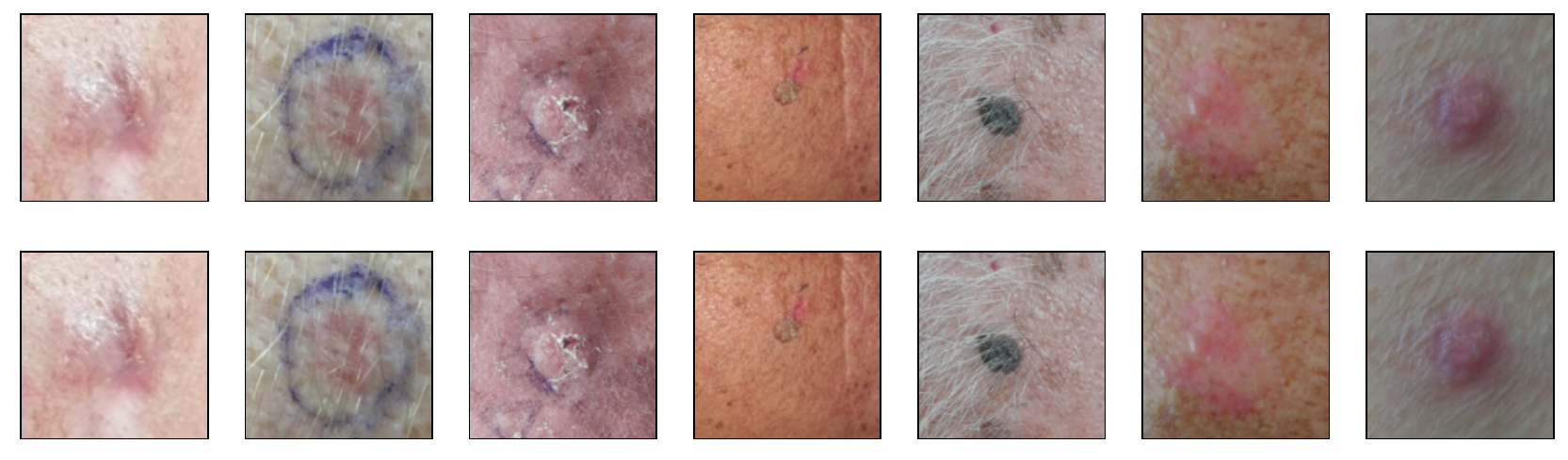}}

    \subfigure[Label Errors]{\label{fig:PAD-UFES-20-Labels}%
      \includegraphics[width=1.0\linewidth]{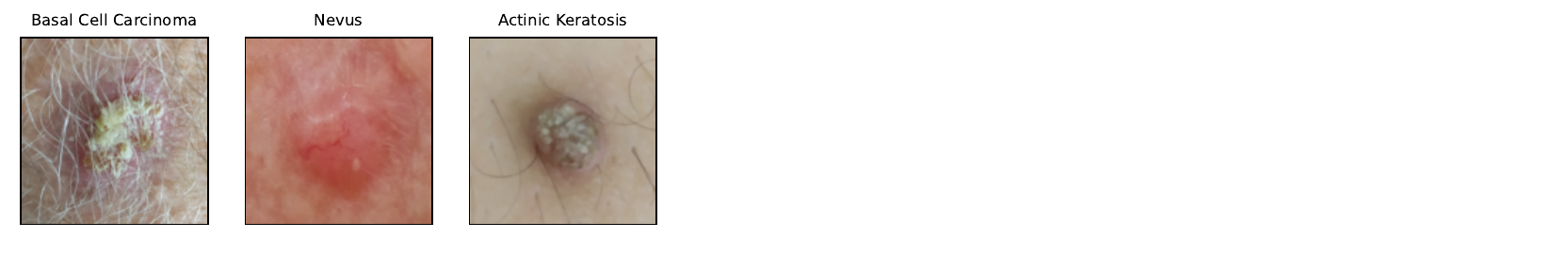}}
  }
\end{figure*}

\begin{figure*}[htbp]
\floatconts
  {fig:SD-128-Issues}
  {\caption{
    Visualization of the data quality issues found in SD-128 by unanimous expert agreement, i.e. 3 irrelevant samples, 157 near duplicates, and 4 label errors.
    Figure \ref{fig:SD-128-Irrelevants} shows the irrelevant samples, \ref{fig:SD-128-Dups} the first 7 near duplicates, and \ref{fig:SD-128-Labels} the label errors.
  }}
  {%
    \subfigure[Irrelevant Samples]{\label{fig:SD-128-Irrelevants}%
      \includegraphics[width=1.0\linewidth]{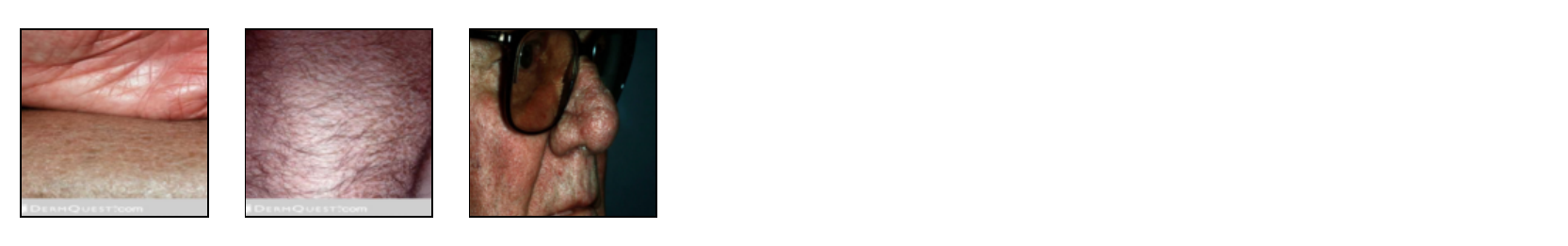}}%
      
    \subfigure[Near Duplicates]{\label{fig:SD-128-Dups}%
      \includegraphics[width=1.0\linewidth]{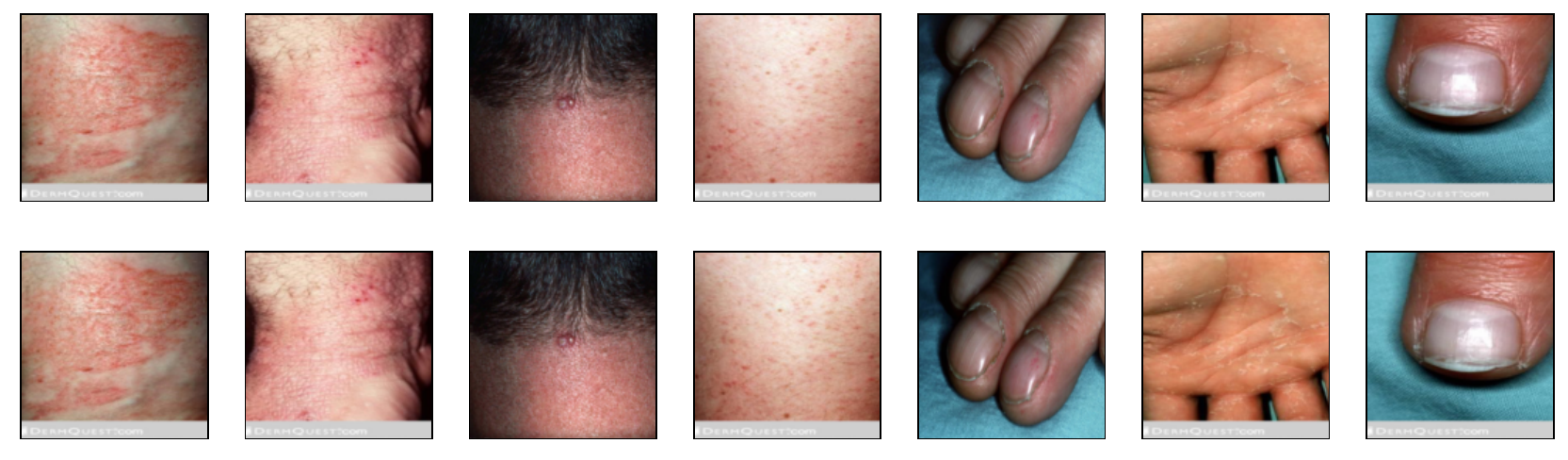}}

    \subfigure[Label Errors]{\label{fig:SD-128-Labels}%
      \includegraphics[width=1.0\linewidth]{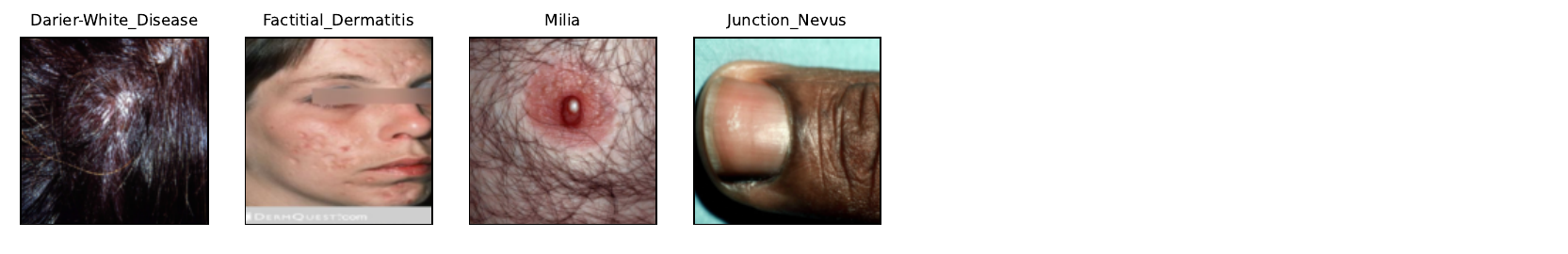}}
  }
\end{figure*}

\section{Detailed data cleaning results}
\label{app:Detailed-Cleaning}

This section provides extended performance results related to dataset cleaning.
Figure \ref{fig:Expert-Verified} visualizes the number of expert-verified samples for each dataset and error type.
Table \ref{tab:Speed-Up-Annotation} shows the speed-up factor for each dataset and noise type when using the proposed resource-efficient data cleaning protocol as opposed to exhaustive annotation.
Table \ref{tab:Quality-Rankings} compares SelfClean with the expert annotations in tabular form.
Figures \ref{fig:Comparison-MedNode-Rankings}, \ref{fig:Comparison-DDI-Rankings}, \ref{fig:Comparison-Derm7pt-Rankings}, \ref{fig:Comparison-PAD-UFES-20-Rankings}, and \ref{fig:Comparison-SD-128-Rankings} illustrate the quality of the candidate rankings by comparing them with each expert's annotations.

\begin{figure*}[htbp]
\floatconts
  {fig:Expert-Verified}
  {\caption{
        Number and percentage of expert-confirmed samples before the annotation was terminated by the stopping criteria for each dataset, expert, and issue type.
  }}
  {%
    \includegraphics[width=1.0\textwidth]{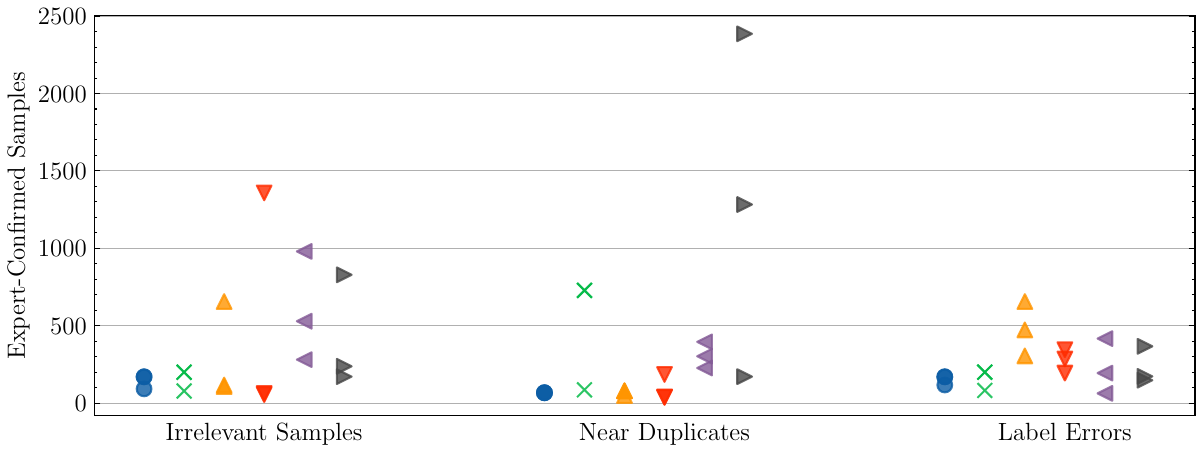}
    \includegraphics[width=1.0\textwidth]{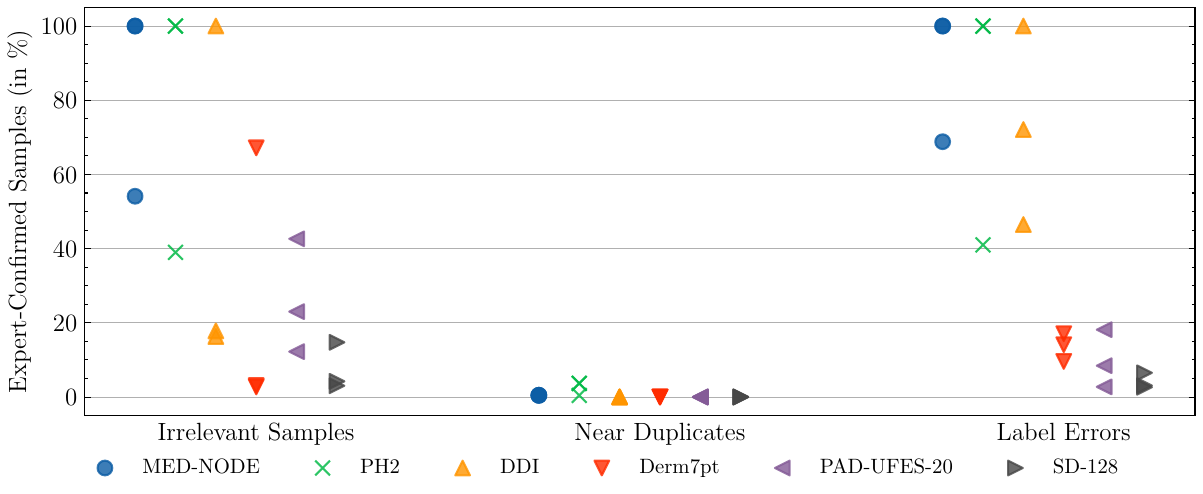}
  }
\end{figure*}

\begin{table*}[htbp]
\centering
\floatconts
  {tab:Quality-Rankings}
  {\caption{
        Comparison of the SelfClean ranking with the labels obtained by unanimous agreement of all expert annotators.
        Note that we only consider samples annotated by all experts.
        For reference, we include the proportion of positive samples, also corresponding to the not-informed baseline performing best in terms of AP.
        PH2 is not evaluated, as it was found to contain no issues.
  }}
  {\begin{tabular}{l l 
        S[table-format=3.2] 
        S[table-format=3.2]
        S[table-format=3.2]
        S[table-format=3.2]
        }
        \toprule
        \bfseries Dataset &
        \bfseries Metadata & 
        \bfseries $\text{Pos. Samples (\%)}$ & 
        \bfseries $\text{AUROC (\%)}$ & 
        \bfseries $\text{AP (\%)}$ &
        \bfseries $\text{AUPRG (\%)}$ \\
        \midrule
        MED-NODE & Irrelevant Samples & 3.3 & 74.2 & 37.8 & 96.6 \\
        MED-NODE & Near Duplicates & 1.5 & 87.9 & 11.1 & 87.9 \\
        MED-NODE & Label Errors & 1.7 & 74.8 & 26.7 & 97.4 \\
        \midrule
        DDI & Irrelevant Samples & 2.8 & 92.3 & 50.8 & 99.5 \\
        DDI & Near Duplicates & 12.2 & 93.8 & 85.3 & 98.9 \\
        DDI & Label Errors & 2.6 & 73.2 & 9.3 & 82.1 \\
        \midrule
        Derm7pt & Irrelevant Samples & 1.8 & 94.4 & 25.0 & 94.4 \\
        Derm7pt & Near Duplicates & 24.3 & 100.0 & 100.0 & 100.0 \\
        Derm7pt & Label Errors & 1.0 & 32.0 & 1.1 & 0.3 \\
        \midrule
        PAD-UFES-20 & Irrelevant Samples & 0.7 & 99.8 & 83.3 & 100.0 \\
        PAD-UFES-20 & Near Duplicates & 24.7 & 82.9 & 67.0 & 86.6 \\
        PAD-UFES-20 & Label Errors & 4.8 & 67.8 & 10.2 & 63.6 \\
        \midrule
        SD-128 & Irrelevant Samples & 1.8 & 22.0 & 1.5 & 0.1 \\
        SD-128 & Near Duplicates & 91.2 & 74.7 & 96.6 & 14.2 \\
        SD-128 & Label Errors & 2.7 & 54.2 & 3.9 & 38.3 \\
        \bottomrule
    \end{tabular}}
\end{table*}

\begin{table*}[htbp]
\centering
\floatconts
  {tab:Speed-Up-Annotation}
  {\caption{
        Speed-up factor of the annotation process with the proposed data cleaning protocol, calculated as the reciprocal of the fraction of annotated samples for each dataset and noise type.
        Additionally the micro and macro-average of each noise type is reported.
  }}
  {\begin{tabular}{l 
        S[table-format=3.2]
        S[table-format=3.2]
        S[table-format=3.2]
        S[table-format=3.2]
        }
        \toprule
        \bfseries Dataset &
        \bfseries $\text{Irrelevant Samples}$ & 
        \bfseries $\text{Near Duplicates}$ & 
        \bfseries $\text{Label Errors}$ \\
        \midrule
        MED-NODE & 1.9 & 214.4 & 1.5 \\
        PH2 & 2.6 & 231.4 & 2.4 \\
        DDI & 6.1 & 4384.5 & 2.2 \\
        Derm7pt & 36.8 & 55222.5 & 10.4 \\
        PAD-UFES-20 & 8.2 & 11626.7 & 36.5 \\
        SD-128 & 32.9 & 92302.8 & 38.0 \\
        \midrule
        Micro-Average & 14.0 & 32520.2 & 12.1 \\
        Macro-Average & 14.8 & 27330.4 & 15.2 \\
        \bottomrule
    \end{tabular}}
\end{table*}

\begin{figure*}[htbp]
\floatconts
  {fig:Comparison-MedNode-Rankings}
  {\caption{
    Comparison between the candidate rankings and expert annotators, as well as their unanimous agreement for MED-NODE. 
    Performance was measured in terms of \gls*{auroc}, \gls*{ap}, and \gls*{auprg}.
    Figure \ref{fig:Comp-MedNode-Irrelevants} shows the comparison for irrelevant samples, \ref{fig:Comp-MedNode-Dups} for near duplicates, and \ref{fig:Comp-MedNode-Labels} for label errors.
  }}
  {%
    \subfigure[Irrelevant Samples]{\label{fig:Comp-MedNode-Irrelevants}%
      \includegraphics[width=1.0\linewidth]{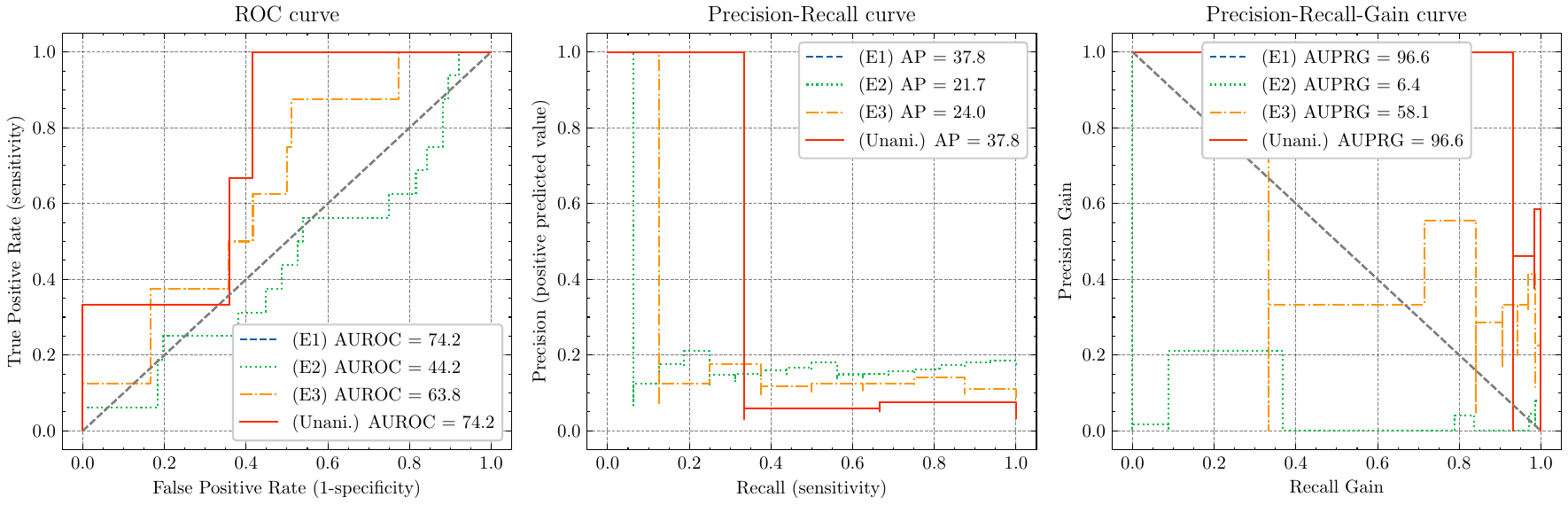}}%
      
    \subfigure[Near Duplicates]{\label{fig:Comp-MedNode-Dups}%
      \includegraphics[width=1.0\linewidth]{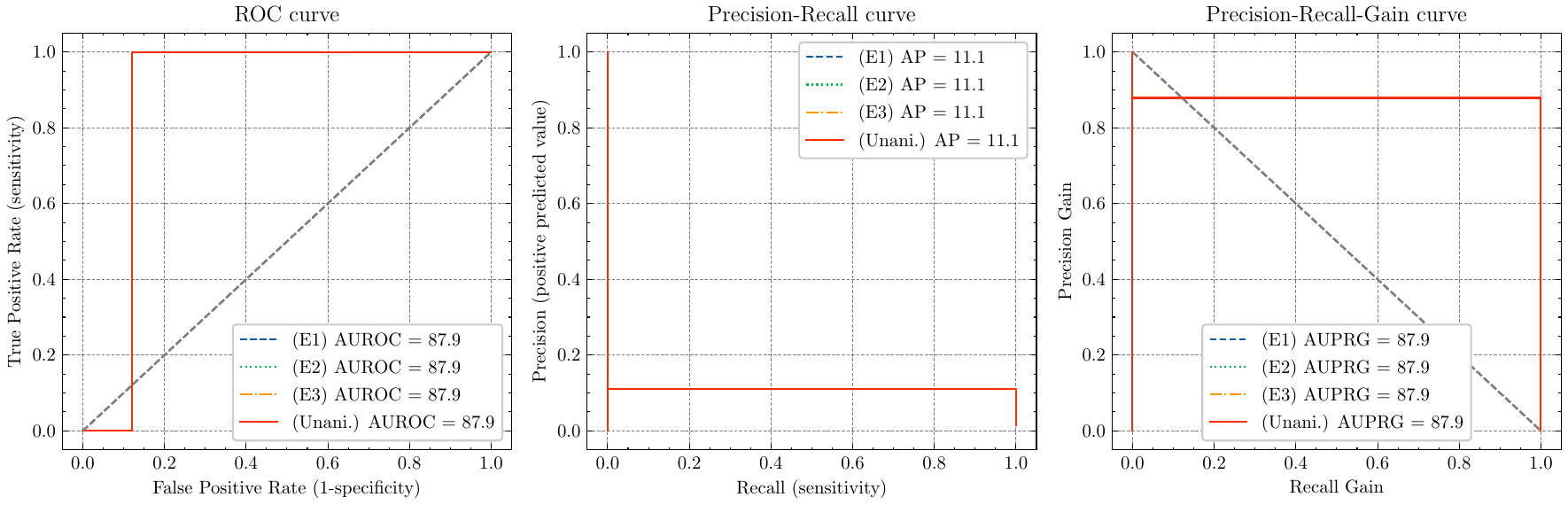}}

    \subfigure[Label Errors]{\label{fig:Comp-MedNode-Labels}%
      \includegraphics[width=1.0\linewidth]{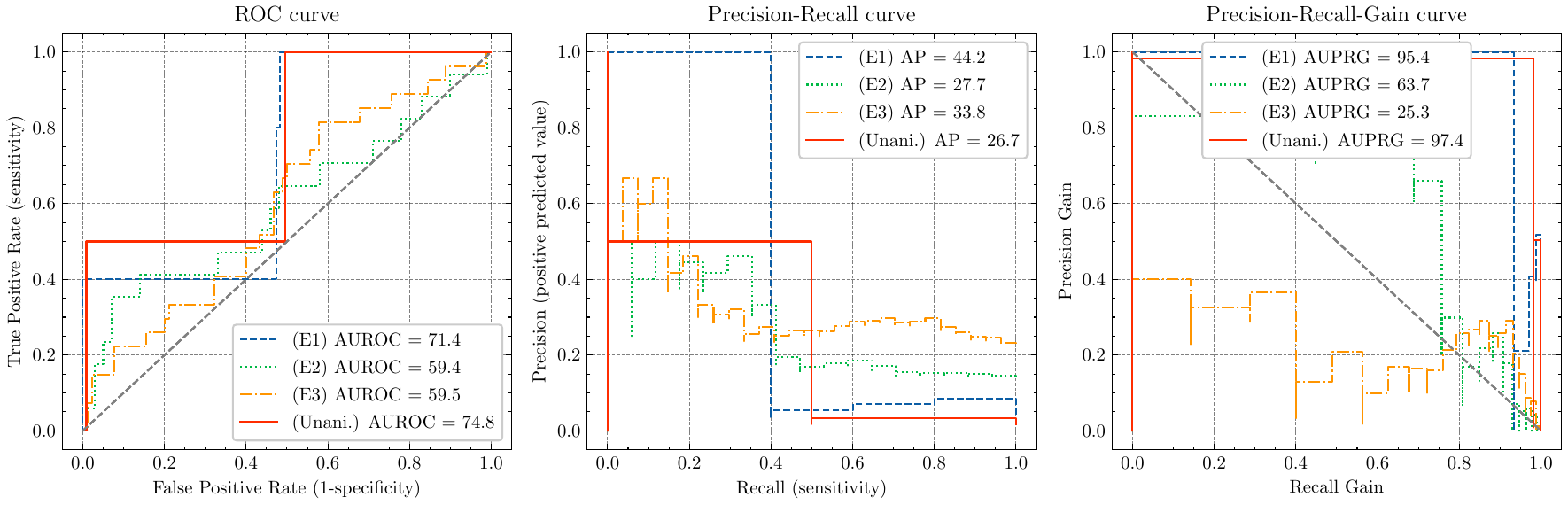}}
  }
\end{figure*}

\begin{figure*}[htbp]
\floatconts
  {fig:Comparison-DDI-Rankings}
  {\caption{
    Comparison between the candidate rankings and expert annotators, as well as their unanimous agreement for DDI. 
    Performance was measured in terms of \gls*{auroc}, \gls*{ap}, and \gls*{auprg}.
    Figure \ref{fig:Comp-DDI-Irrelevants} shows the comparison for irrelevant samples, \ref{fig:Comp-DDI-Dups} for near duplicates, and \ref{fig:Comp-DDI-Labels} for label errors.
  }}
  {%
    \subfigure[Irrelevant Samples]{\label{fig:Comp-DDI-Irrelevants}%
      \includegraphics[width=1.0\linewidth]{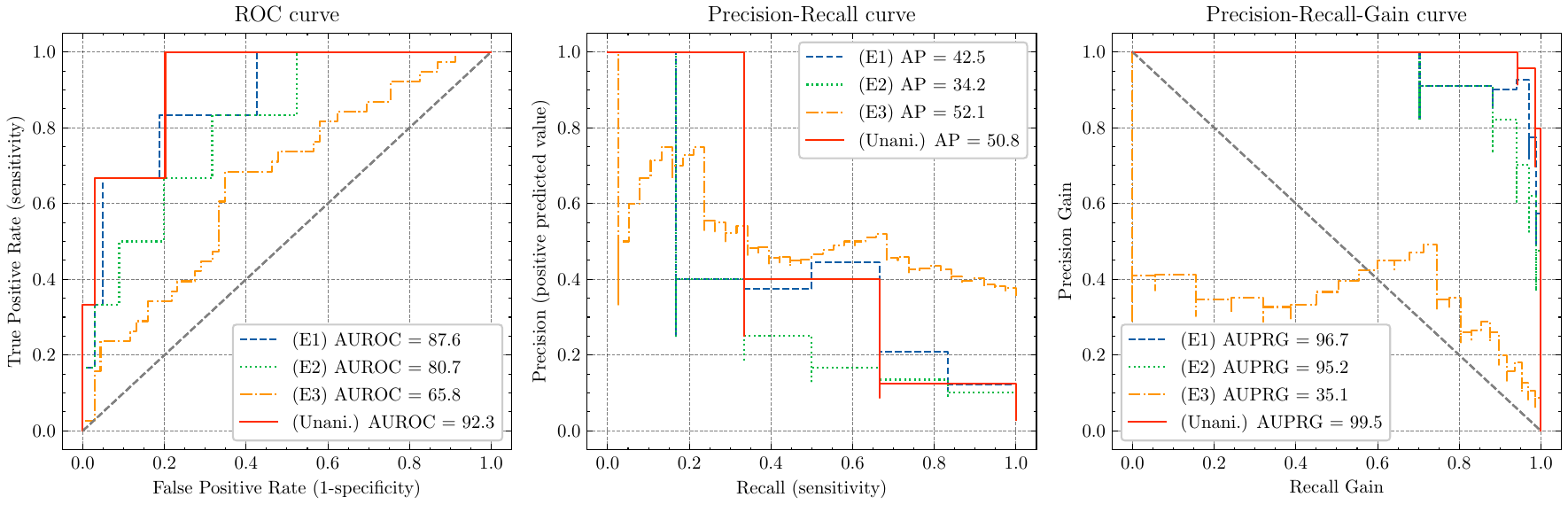}}%
      
    \subfigure[Near Duplicates]{\label{fig:Comp-DDI-Dups}%
      \includegraphics[width=1.0\linewidth]{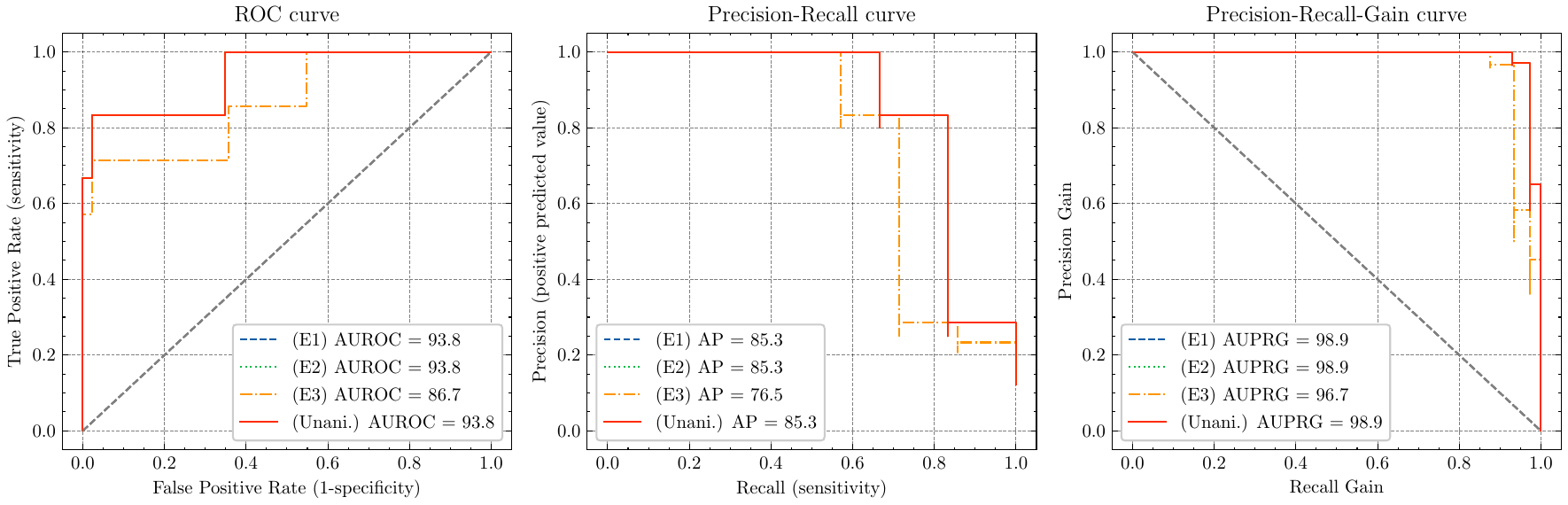}}

    \subfigure[Label Errors]{\label{fig:Comp-DDI-Labels}%
      \includegraphics[width=1.0\linewidth]{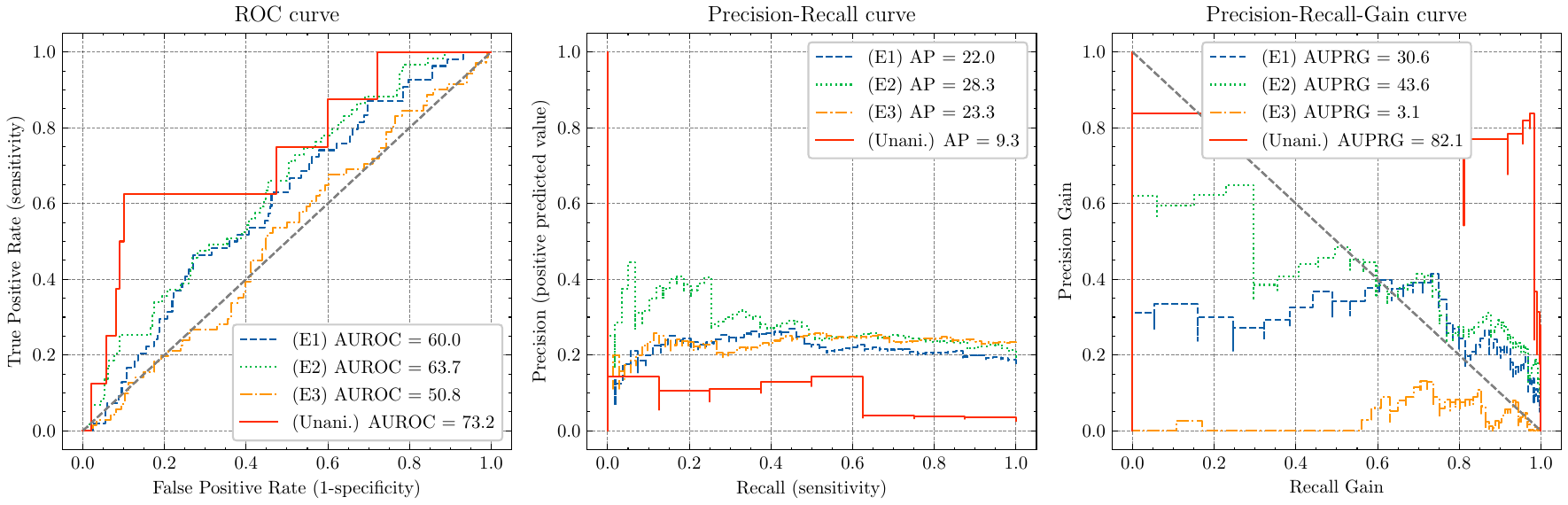}}
  }
\end{figure*}

\begin{figure*}[htbp]
\floatconts
  {fig:Comparison-Derm7pt-Rankings}
  {\caption{
    Comparison between the candidate rankings and expert annotators, as well as their unanimous agreement for Derm7pt. 
    Performance was measured in terms of \gls*{auroc}, \gls*{ap}, and \gls*{auprg}.
    Figure \ref{fig:Comp-Derm7pt-Irrelevants} shows the comparison for irrelevant samples, \ref{fig:Comp-Derm7pt-Dups} for near duplicates, and \ref{fig:Comp-Derm7pt-Labels} for label errors.
  }}
  {%
    \subfigure[Irrelevant Samples]{\label{fig:Comp-Derm7pt-Irrelevants}%
      \includegraphics[width=1.0\linewidth]{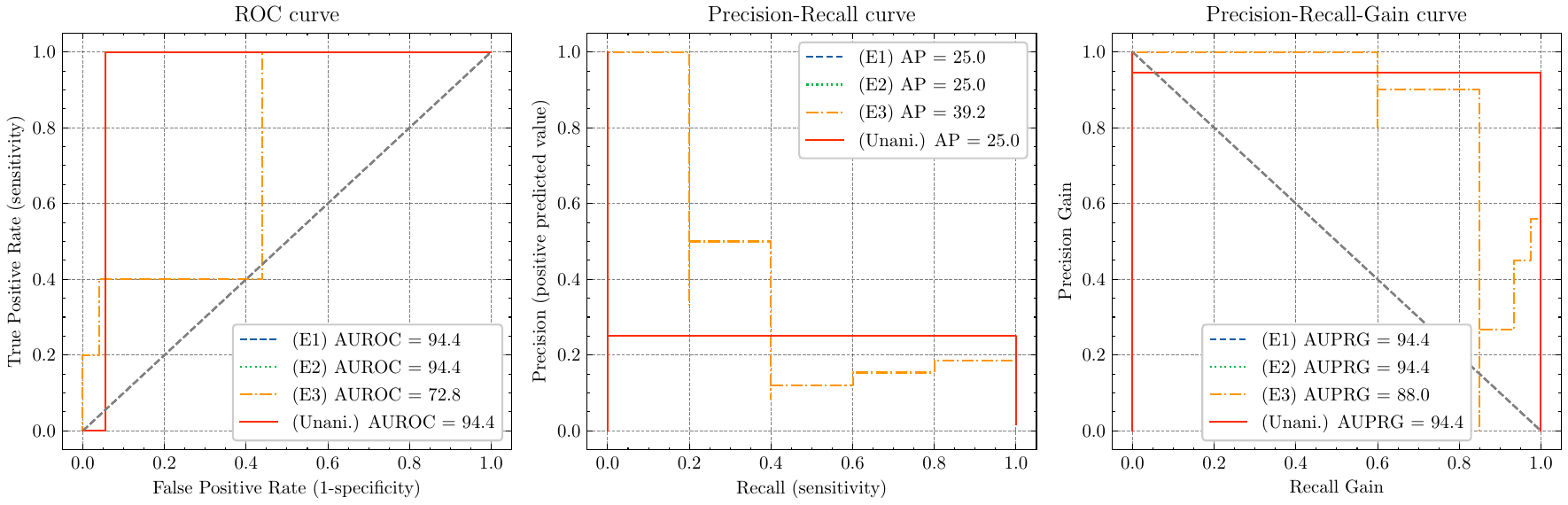}}%
      
    \subfigure[Near Duplicates]{\label{fig:Comp-Derm7pt-Dups}%
      \includegraphics[width=1.0\linewidth]{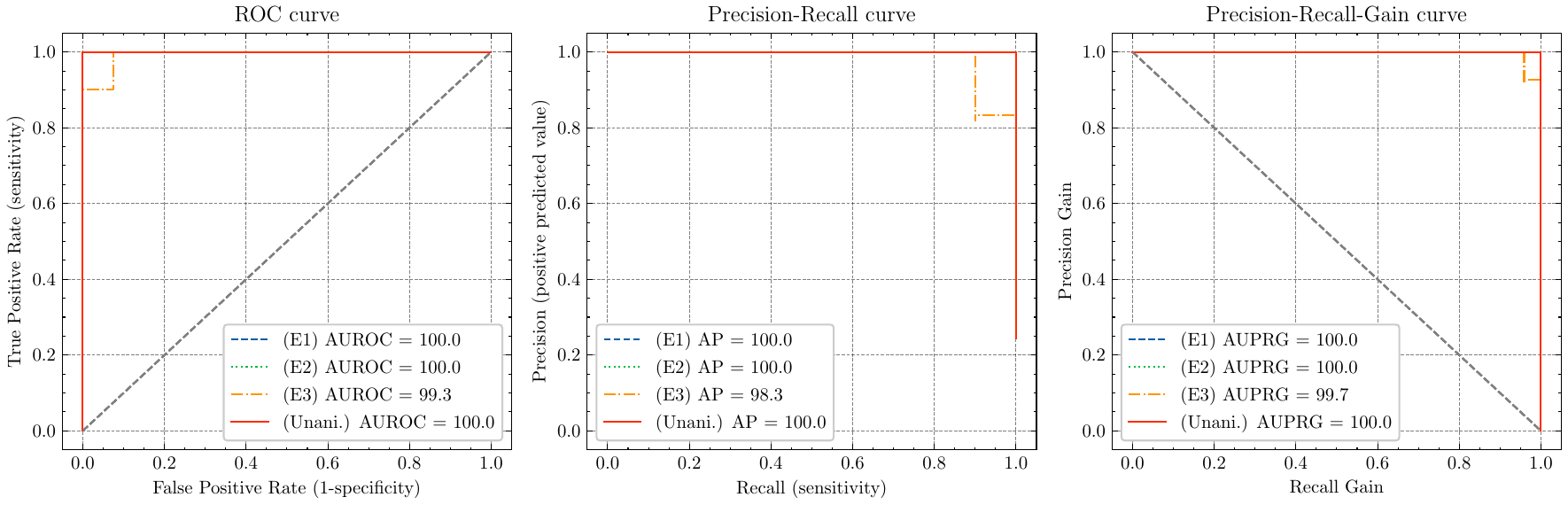}}

    \subfigure[Label Errors]{\label{fig:Comp-Derm7pt-Labels}%
      \includegraphics[width=1.0\linewidth]{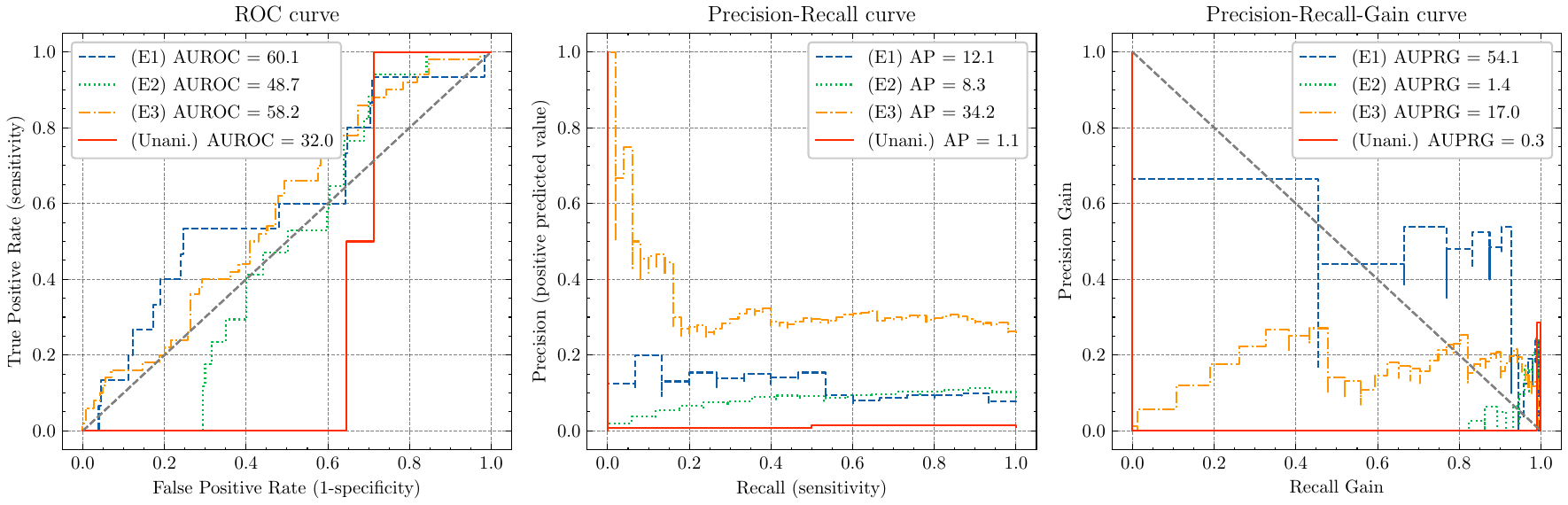}}
  }
\end{figure*}

\begin{figure*}[htbp]
\floatconts
  {fig:Comparison-PAD-UFES-20-Rankings}
  {\caption{
    Comparison between the candidate rankings and expert annotators, as well as their unanimous agreement for PAD-UFES-20. 
    Performance was measured in terms of \gls*{auroc}, \gls*{ap}, and \gls*{auprg}.
    Figure \ref{fig:Comp-PAD-UFES-20-Irrelevants} shows the comparison for irrelevant samples, \ref{fig:Comp-PAD-UFES-20-Dups} for near duplicates, and \ref{fig:Comp-PAD-UFES-20-Labels} for label errors.
  }}
  {%
    \subfigure[Irrelevant Samples]{\label{fig:Comp-PAD-UFES-20-Irrelevants}%
      \includegraphics[width=1.0\linewidth]{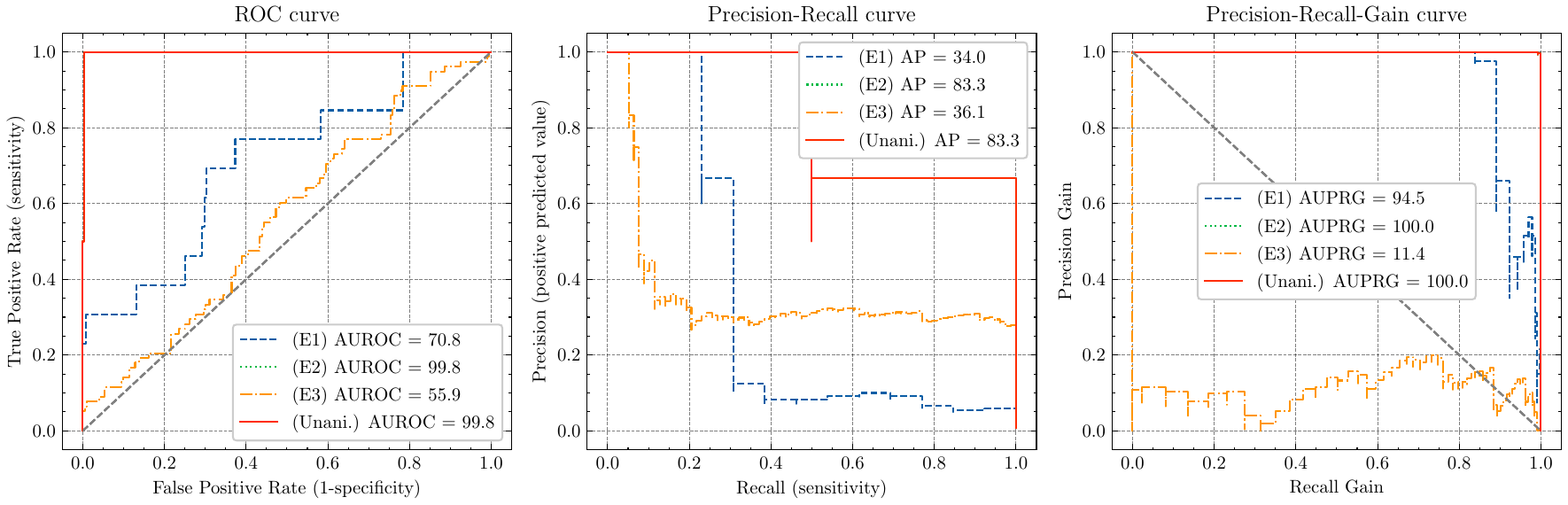}}%
      
    \subfigure[Near Duplicates]{\label{fig:Comp-PAD-UFES-20-Dups}%
      \includegraphics[width=1.0\linewidth]{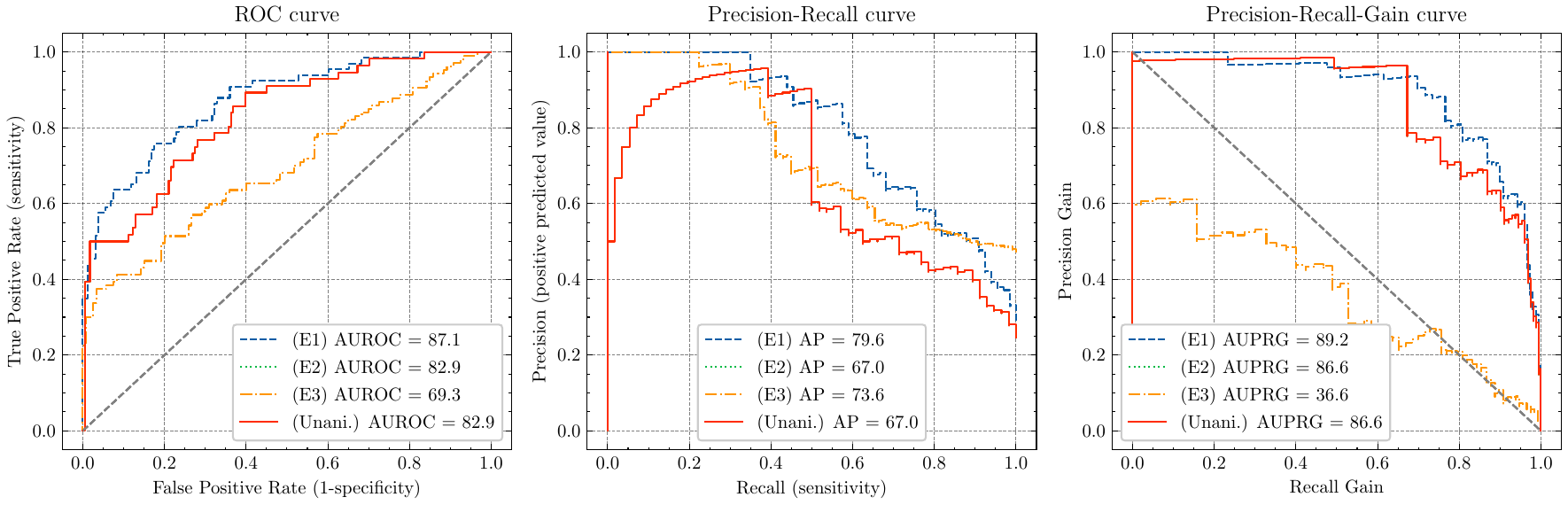}}

    \subfigure[Label Errors]{\label{fig:Comp-PAD-UFES-20-Labels}%
      \includegraphics[width=1.0\linewidth]{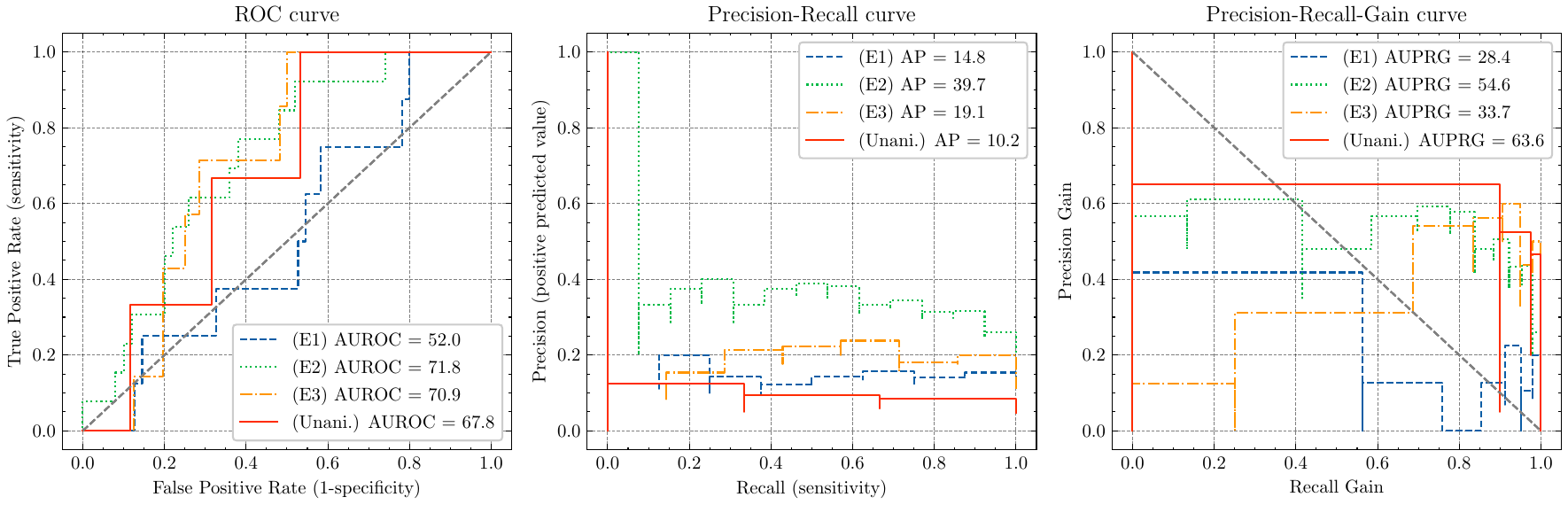}}
  }
\end{figure*}

\begin{figure*}[htbp]
\floatconts
  {fig:Comparison-SD-128-Rankings}
  {\caption{
    Comparison between the candidate rankings and expert annotators, as well as their unanimous agreement for SD-128. 
    Performance was measured in terms of \gls*{auroc}, \gls*{ap}, and \gls*{auprg}.
    Figure \ref{fig:Comp-SD-128-Irrelevants} shows the comparison for irrelevant samples, \ref{fig:Comp-SD-128-Dups} for near duplicates, and \ref{fig:Comp-SD-128-Labels} for label errors.
  }}
  {%
    \subfigure[Irrelevant Samples]{\label{fig:Comp-SD-128-Irrelevants}%
      \includegraphics[width=1.0\linewidth]{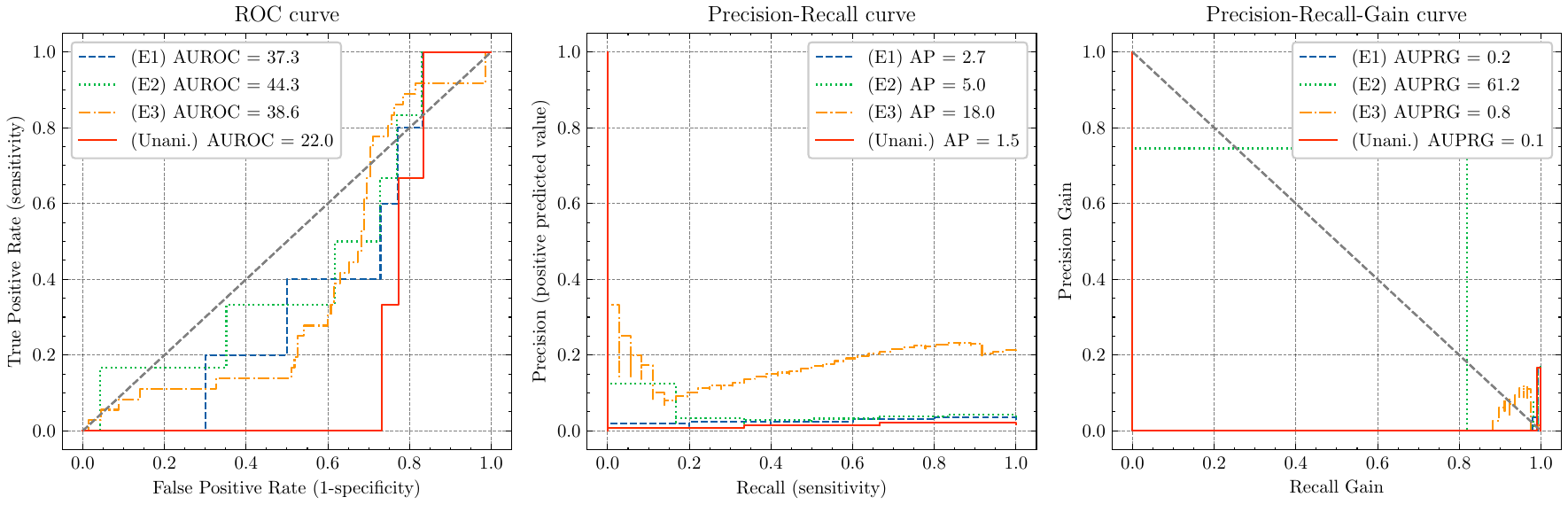}}%
      
    \subfigure[Near Duplicates]{\label{fig:Comp-SD-128-Dups}%
      \includegraphics[width=1.0\linewidth]{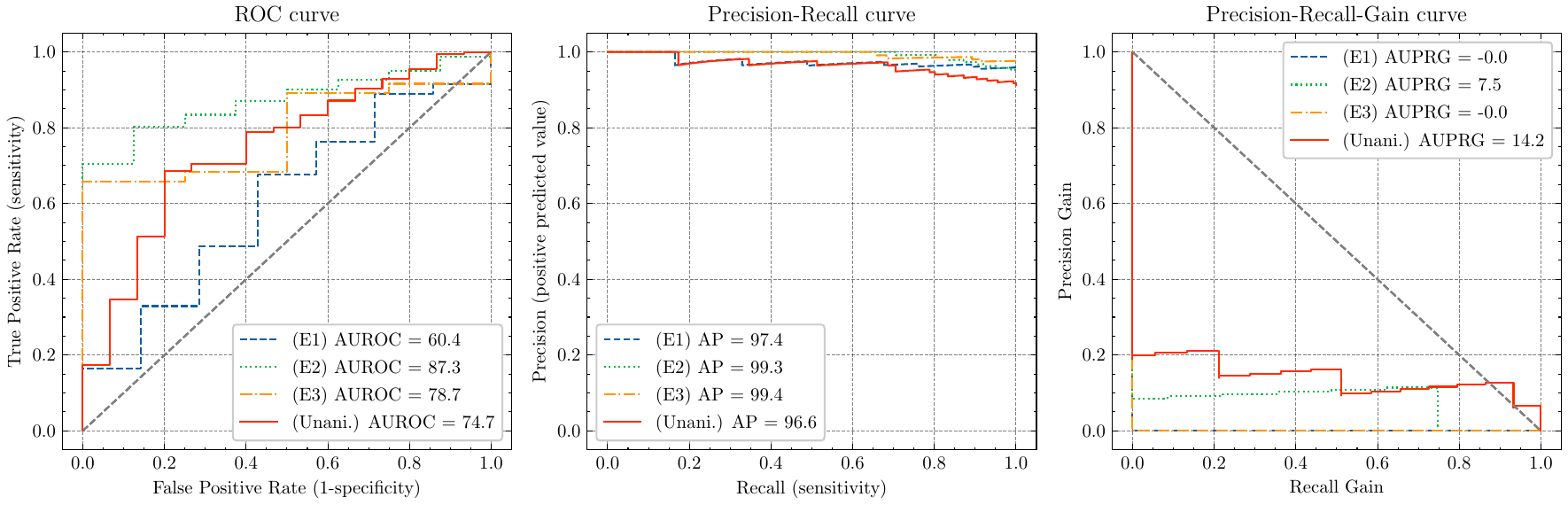}}

    \subfigure[Label Errors]{\label{fig:Comp-SD-128-Labels}%
      \includegraphics[width=1.0\linewidth]{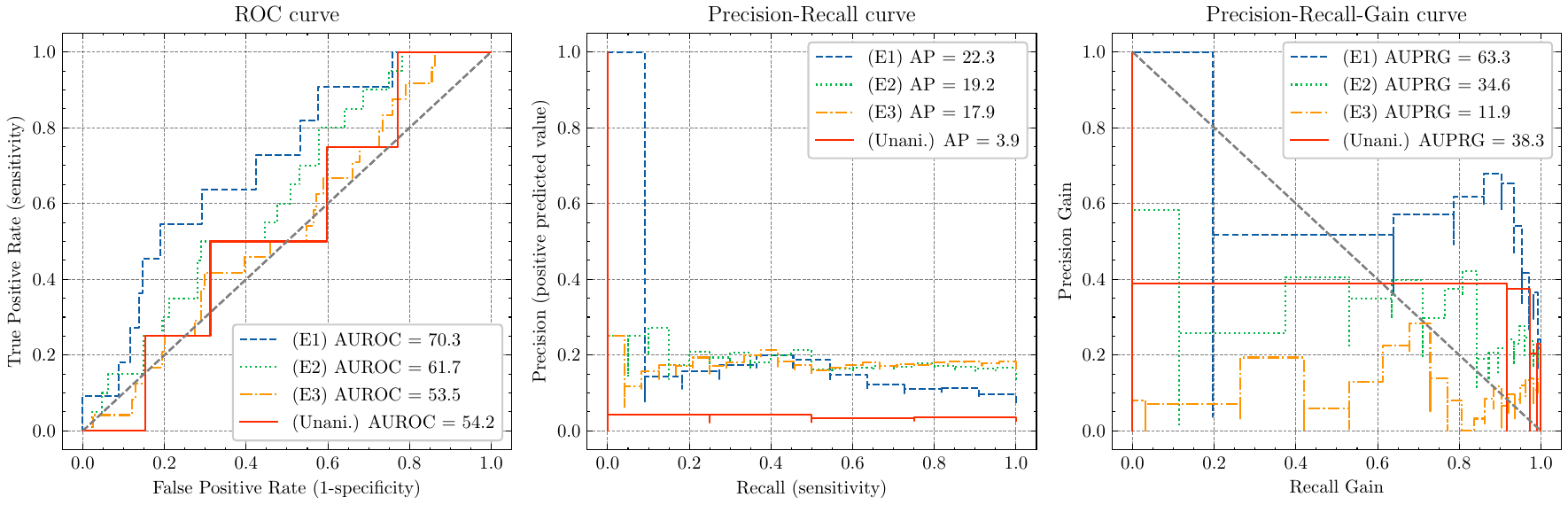}}
  }
\end{figure*}

\section{Ablation stopping criteria} \label{app:Ablation-Stopping}

\begin{figure}[htbp]
\floatconts
  {fig:Stopping-Sensitivity-MED-NODE}
  {\caption{
    Dependence of the number of detected data quality issues on the parameters $p_{\text{chance}}$ and $p_+$ of the stopping criteria for MED-NODE.
    Both parameters start at the default value used throughout the paper $p_\text{chance} = p_+ = 0.05$ and are increased one at a time to $p_\text{chance} = p_+ = 1.0$.
    Figure \ref{fig:Stopping-Sensitivity-MED-NODE-Irrelevants} shows the behavior for irrelevant samples, \ref{fig:Stopping-Sensitivity-MED-NODE-Dups} for near duplicates, and \ref{fig:Stopping-Sensitivity-MED-NODE-Labels} for label errors.
  }}
  {%
    \subfigure[Irrelevant Samples]{\label{fig:Stopping-Sensitivity-MED-NODE-Irrelevants}%
      \includegraphics[width=1.0\linewidth]{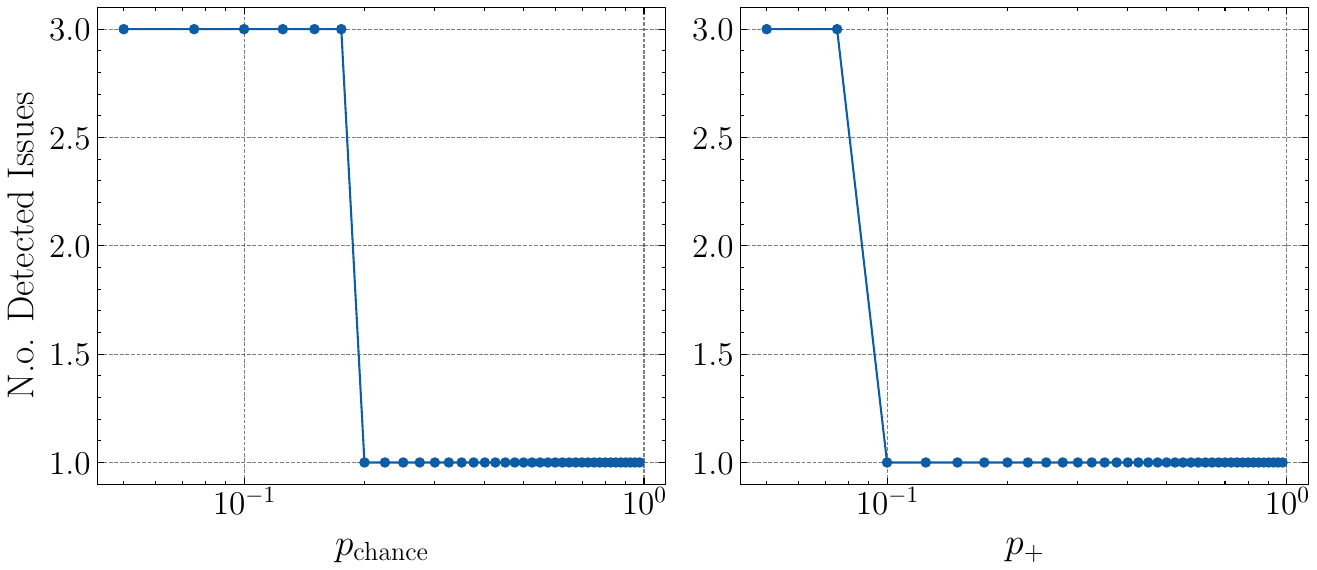}}%
      
    \subfigure[Near Duplicates]{\label{fig:Stopping-Sensitivity-MED-NODE-Dups}%
      \includegraphics[width=1.0\linewidth]{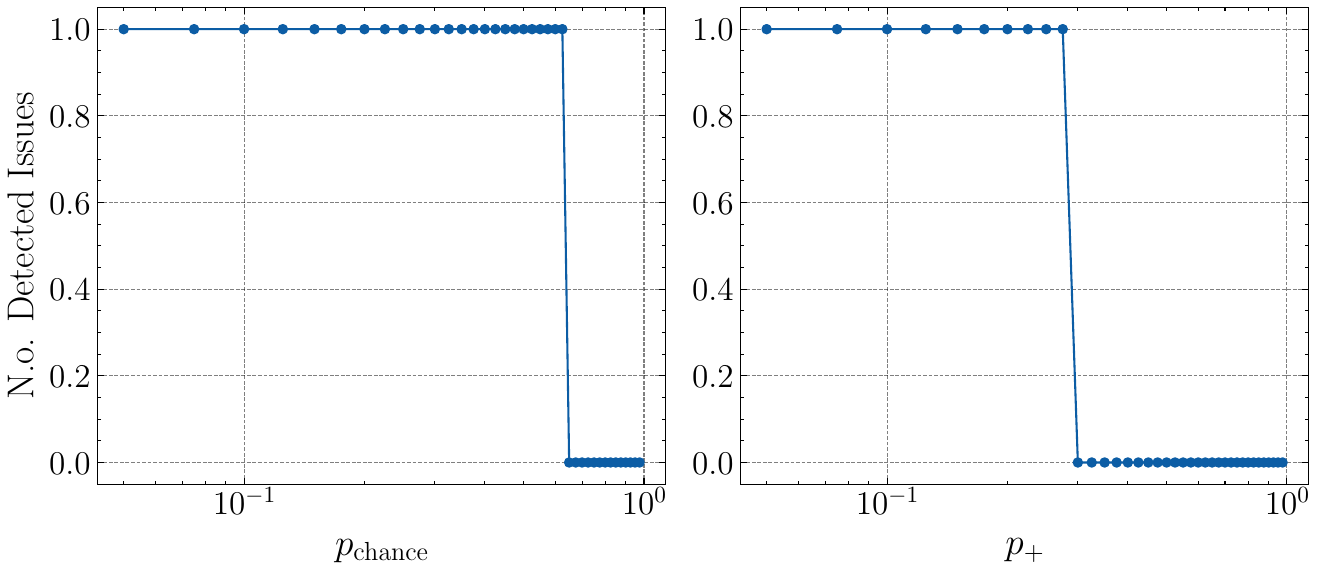}}

    \subfigure[Label Errors]{\label{fig:Stopping-Sensitivity-MED-NODE-Labels}%
      \includegraphics[width=1.0\linewidth]{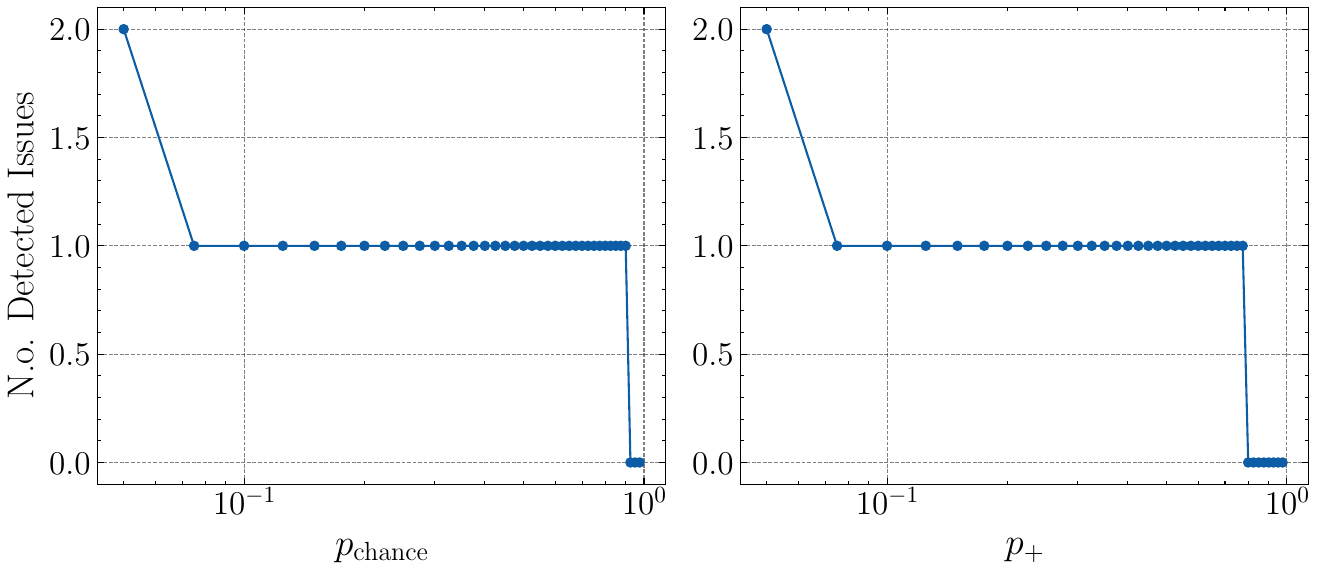}}
  }
\end{figure}

\begin{figure}[htbp]
\floatconts
  {fig:Stopping-Sensitivity-Derm7pt}
  {\caption{
    Dependence of the number of detected data quality issues on the parameters $p_{\text{chance}}$ and $p_+$ of the stopping criteria for Derm7pt.
    Both parameters start at the default value used throughout the paper $p_\text{chance} = p_+ = 0.05$ and are increased one at a time to $p_\text{chance} = p_+ = 1.0$.
    Figure \ref{fig:Stopping-Sensitivity-Derm7pt-Irrelevants} shows the behavior for irrelevant samples, \ref{fig:Stopping-Sensitivity-Derm7pt-Dups} for near duplicates, and \ref{fig:Stopping-Sensitivity-Derm7pt-Labels} for label errors.
  }}
  {%
    \subfigure[Irrelevant Samples]{\label{fig:Stopping-Sensitivity-Derm7pt-Irrelevants}%
      \includegraphics[width=1.0\linewidth]{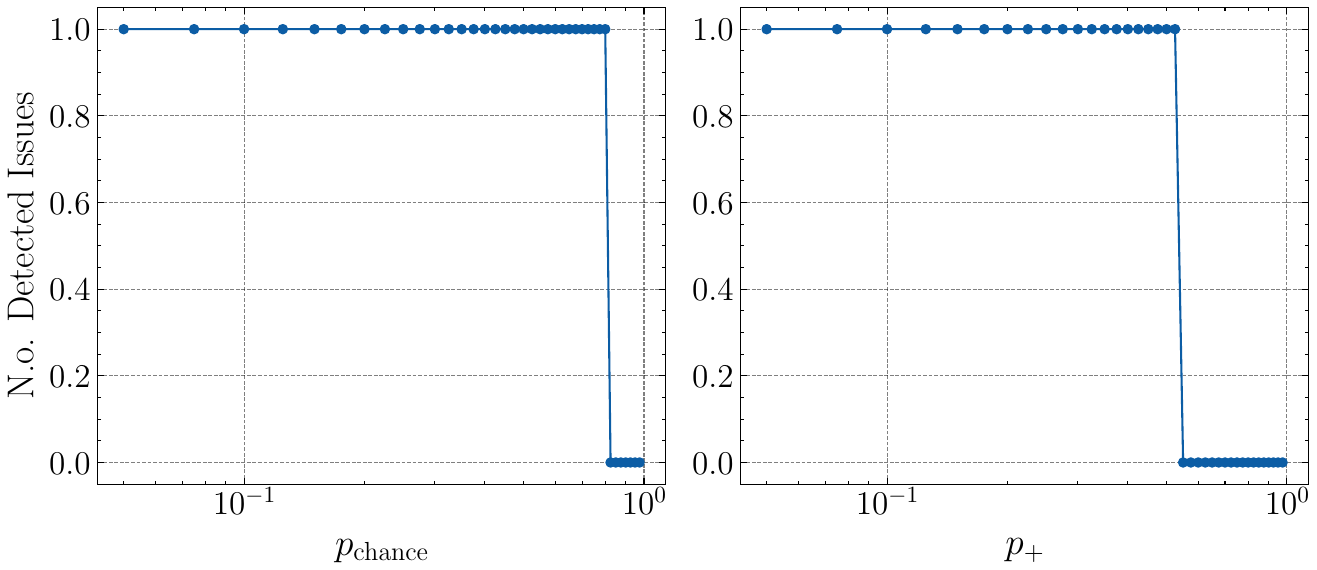}}%
      
    \subfigure[Near Duplicates]{\label{fig:Stopping-Sensitivity-Derm7pt-Dups}%
      \includegraphics[width=1.0\linewidth]{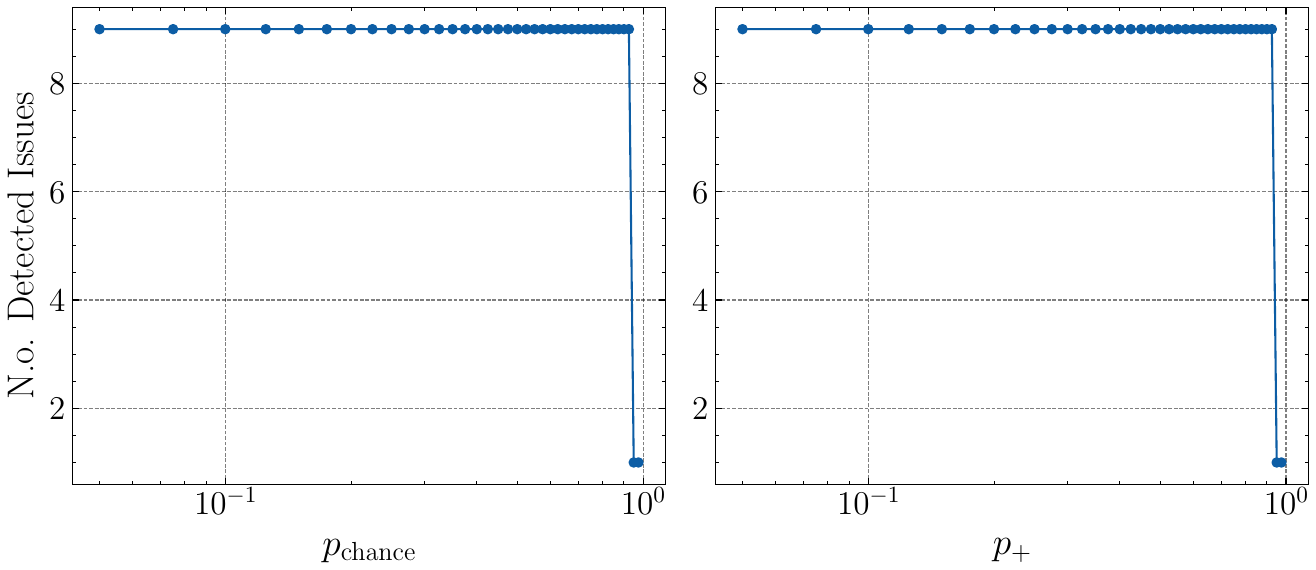}}

    \subfigure[Label Errors]{\label{fig:Stopping-Sensitivity-Derm7pt-Labels}%
      \includegraphics[width=1.0\linewidth]{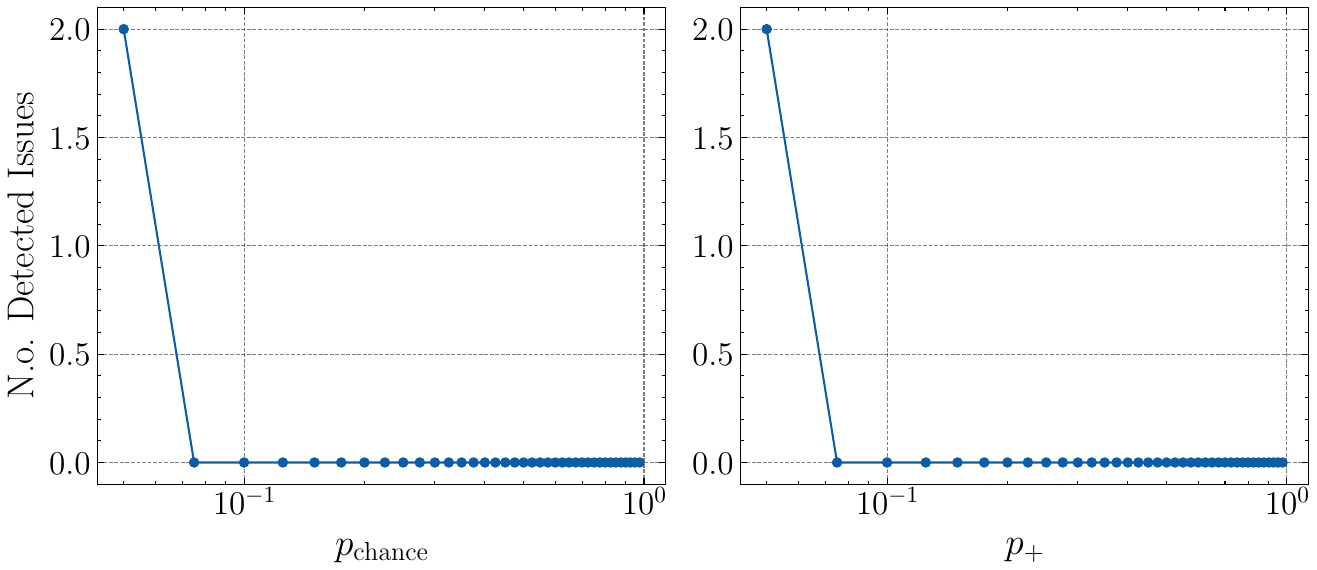}}
  }
\end{figure}

\begin{figure}[htbp]
\floatconts
  {fig:Stopping-Sensitivity-SD-128}
  {\caption{
    Dependence of the number of detected data quality issues on the parameters $p_{\text{chance}}$ and $p_+$ of the stopping criteria for SD-128.
    Both parameters start at the default value used throughout the paper $p_\text{chance} = p_+ = 0.05$ and are increased one at a time to $p_\text{chance} = p_+ = 1.0$.
    Figure \ref{fig:Stopping-Sensitivity-SD-128-Irrelevants} shows the behavior for irrelevant samples, \ref{fig:Stopping-Sensitivity-SD-128-Dups} for near duplicates, and \ref{fig:Stopping-Sensitivity-SD-128-Labels} for label errors.
  }}
  {%
    \subfigure[Irrelevant Samples]{\label{fig:Stopping-Sensitivity-SD-128-Irrelevants}%
      \includegraphics[width=1.0\linewidth]{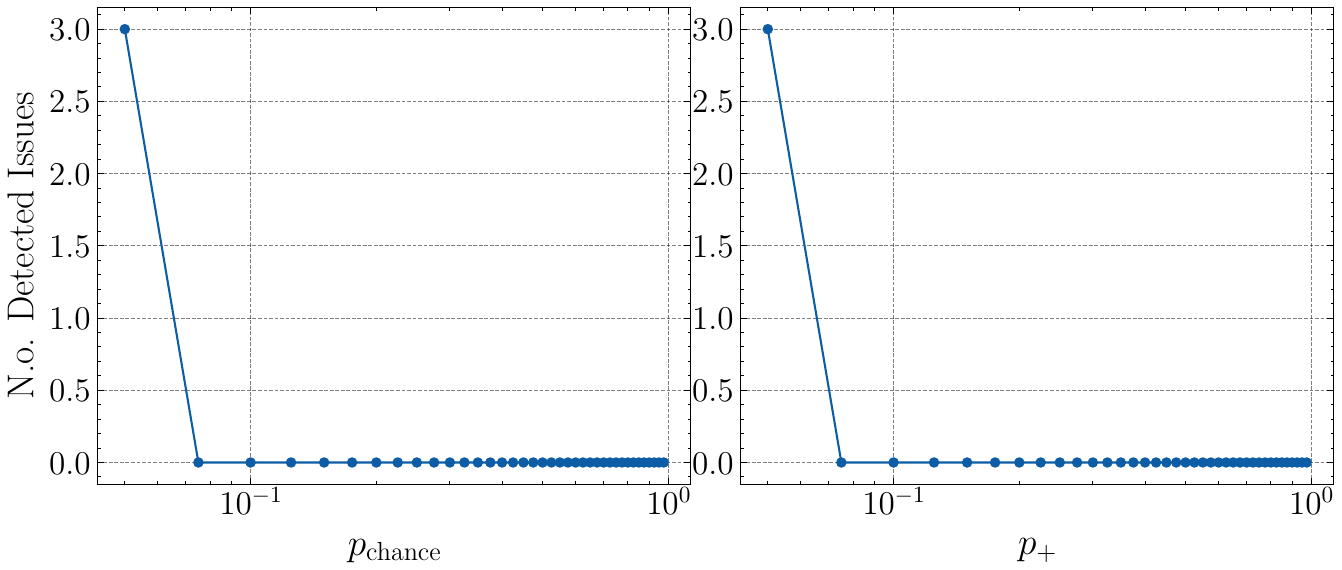}}%
      
    \subfigure[Near Duplicates]{\label{fig:Stopping-Sensitivity-SD-128-Dups}%
      \includegraphics[width=1.0\linewidth]{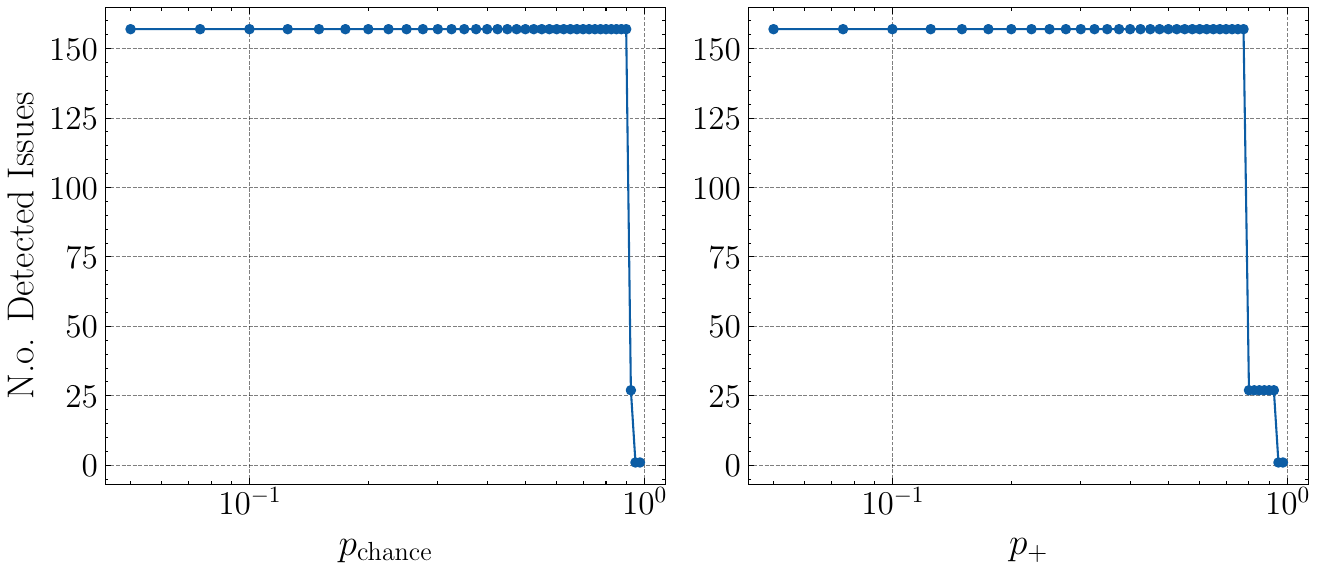}}

    \subfigure[Label Errors]{\label{fig:Stopping-Sensitivity-SD-128-Labels}%
      \includegraphics[width=1.0\linewidth]{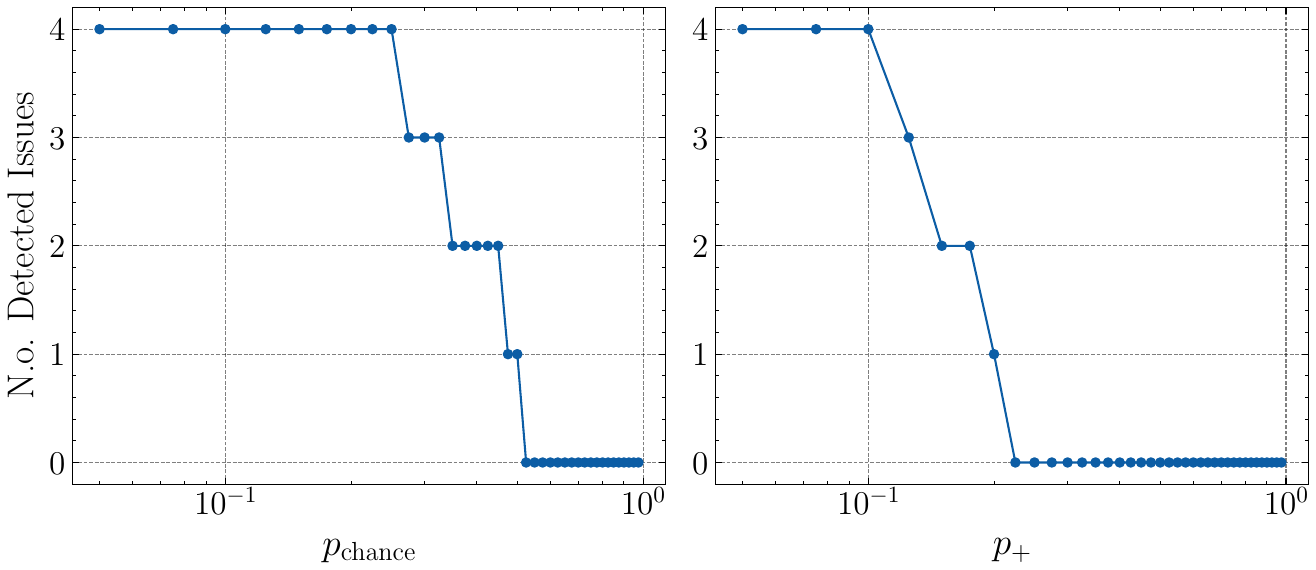}}
  }
\end{figure}

Figure \ref{fig:Stopping-Sensitivity-MED-NODE}, \ref{fig:Stopping-Sensitivity-DDI}, \ref{fig:Stopping-Sensitivity-Derm7pt}, \ref{fig:Stopping-Sensitivity-PAD-UFES-20}, and \ref{fig:Stopping-Sensitivity-SD-128} visualize the dependence of the number of detected samples on the parameters $p_{\text{chance}}$ and $p_+$ of the stopping criteria for each dataset.
We see that results are robust to the choice of parameters as the number of detected issues does not change significantly when increasing $p_{\text{chance}}$ and $p_+$ for most datasets and quality issues.
However, there are a few cases where changing the parameters can lead to detecting fewer problematic instances, such as Figure \ref{fig:Stopping-Sensitivity-MED-NODE-Labels}, \ref{fig:Stopping-Sensitivity-Derm7pt-Labels}, and \ref{fig:Stopping-Sensitivity-SD-128-Irrelevants}, indicating that some data quality issues are found after a larger consecutive number of non-problematic samples.
In summary the results show that, when we stop annotation even earlier, in most cases the number of detected issues is still insensitive to the change of parameters.

\begin{figure}[htbp]
\floatconts
  {fig:Stopping-Sensitivity-DDI}
  {\caption{
    Dependence of the number of detected data quality issues on the parameters $p_{\text{chance}}$ and $p_+$ of the stopping criteria for DDI.
    Both parameters start at the default value used throughout the paper $p_\text{chance} = p_+ = 0.05$ and are increased one at a time to $p_\text{chance} = p_+ = 1.0$.
    Figure \ref{fig:Stopping-Sensitivity-DDI-Irrelevants} shows the behavior for irrelevant samples, \ref{fig:Stopping-Sensitivity-DDI-Dups} for near duplicates, and \ref{fig:Stopping-Sensitivity-DDI-Labels} for label errors.
  }}
  {%
    \subfigure[Irrelevant Samples]{\label{fig:Stopping-Sensitivity-DDI-Irrelevants}%
      \includegraphics[width=1.0\linewidth]{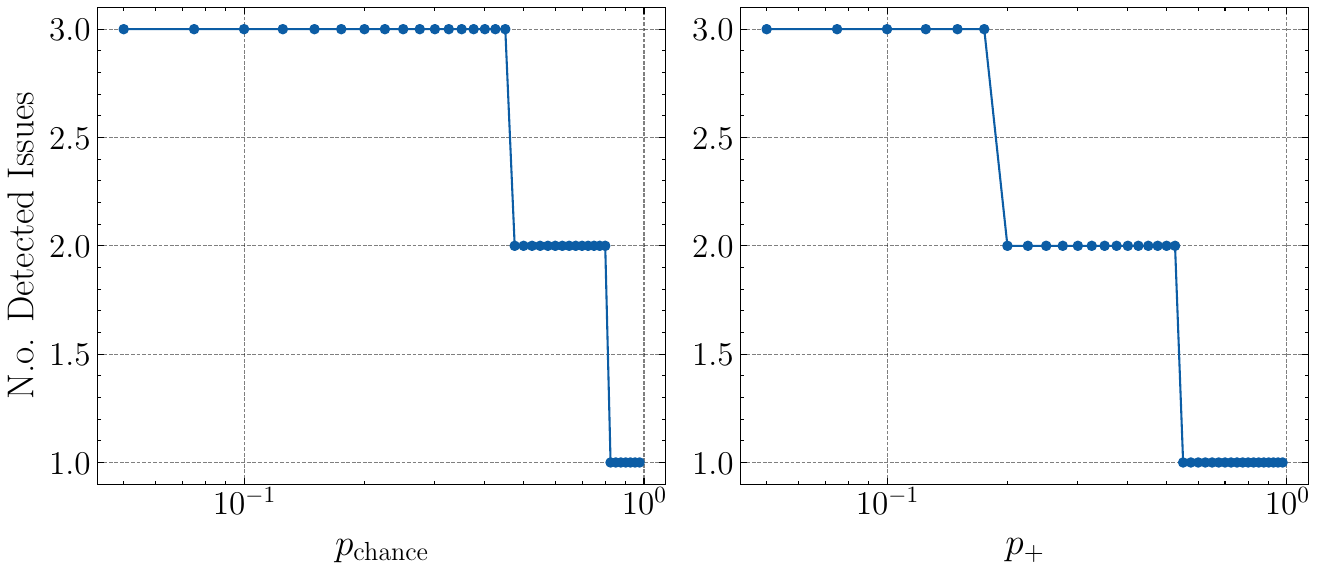}}%
      
    \subfigure[Near Duplicates]{\label{fig:Stopping-Sensitivity-DDI-Dups}%
      \includegraphics[width=1.0\linewidth]{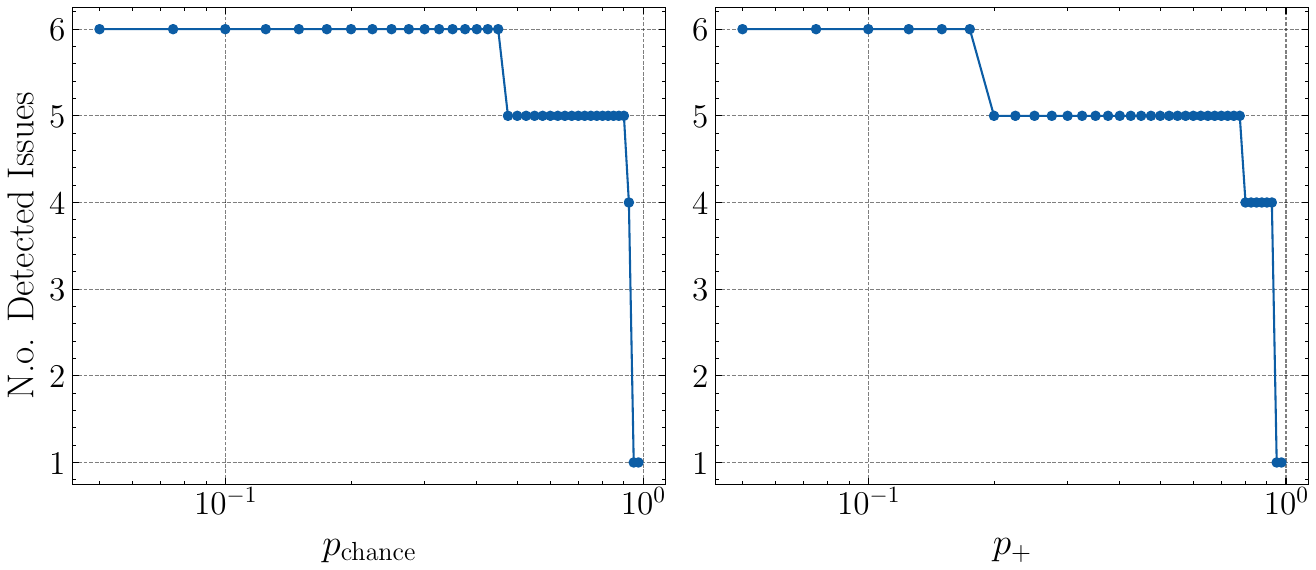}}

    \subfigure[Label Errors]{\label{fig:Stopping-Sensitivity-DDI-Labels}%
      \includegraphics[width=1.0\linewidth]{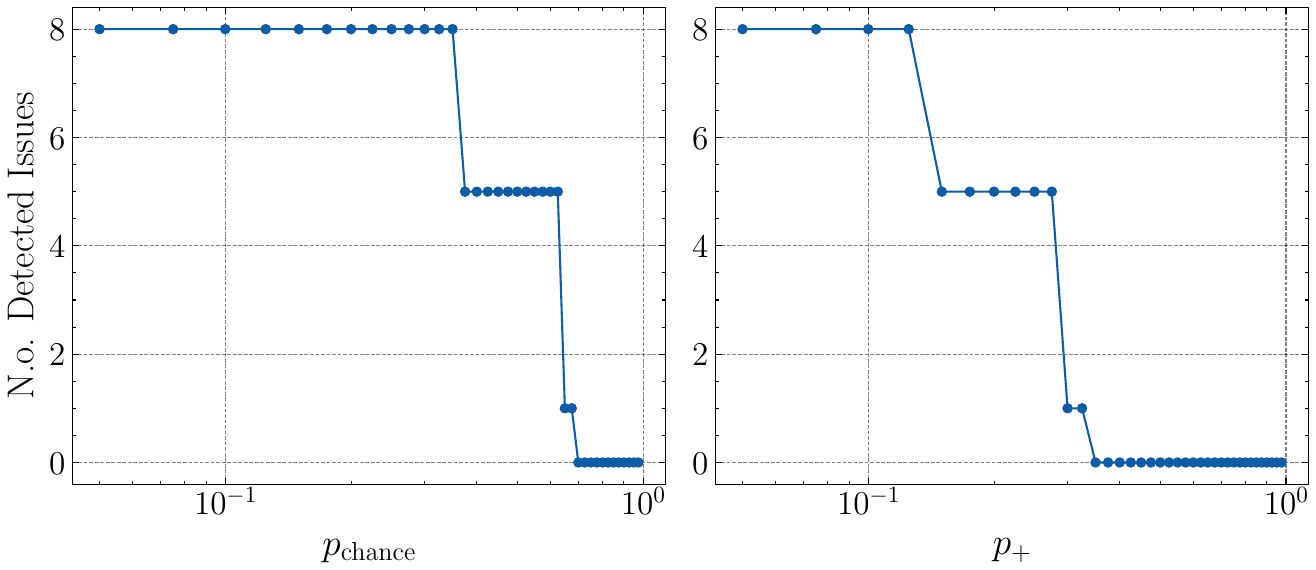}}
  }
\end{figure}

\begin{figure}[htbp]
\floatconts
  {fig:Stopping-Sensitivity-PAD-UFES-20}
  {\caption{
    Dependence of the number of detected data quality issues on the parameters $p_{\text{chance}}$ and $p_+$ of the stopping criteria for PAD-UFES-20.
    Both parameters start at the default value used throughout the paper $p_\text{chance} = p_+ = 0.05$ and are increased one at a time to $p_\text{chance} = p_+ = 1.0$.
    Figure \ref{fig:Stopping-Sensitivity-PAD-UFES-20-Irrelevants} shows the behavior for irrelevant samples, \ref{fig:Stopping-Sensitivity-PAD-UFES-20-Dups} for near duplicates, and \ref{fig:Stopping-Sensitivity-PAD-UFES-20-Labels} for label errors.
  }}
  {%
    \subfigure[Irrelevant Samples]{\label{fig:Stopping-Sensitivity-PAD-UFES-20-Irrelevants}%
      \includegraphics[width=1.0\linewidth]{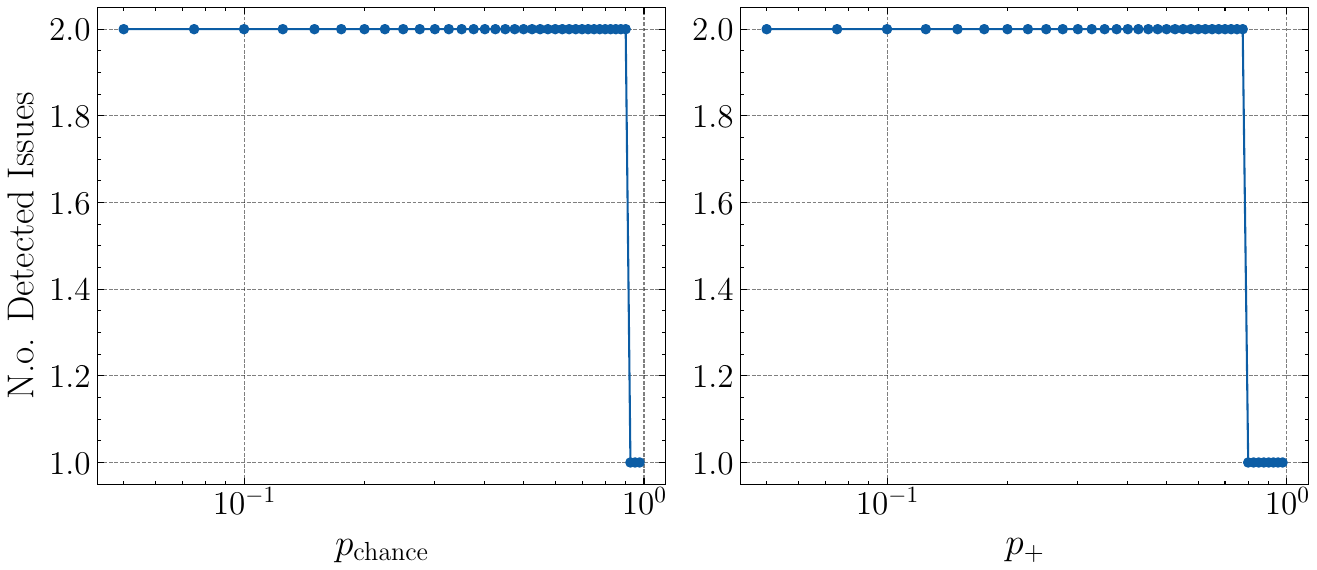}}%
      
    \subfigure[Near Duplicates]{\label{fig:Stopping-Sensitivity-PAD-UFES-20-Dups}%
      \includegraphics[width=1.0\linewidth]{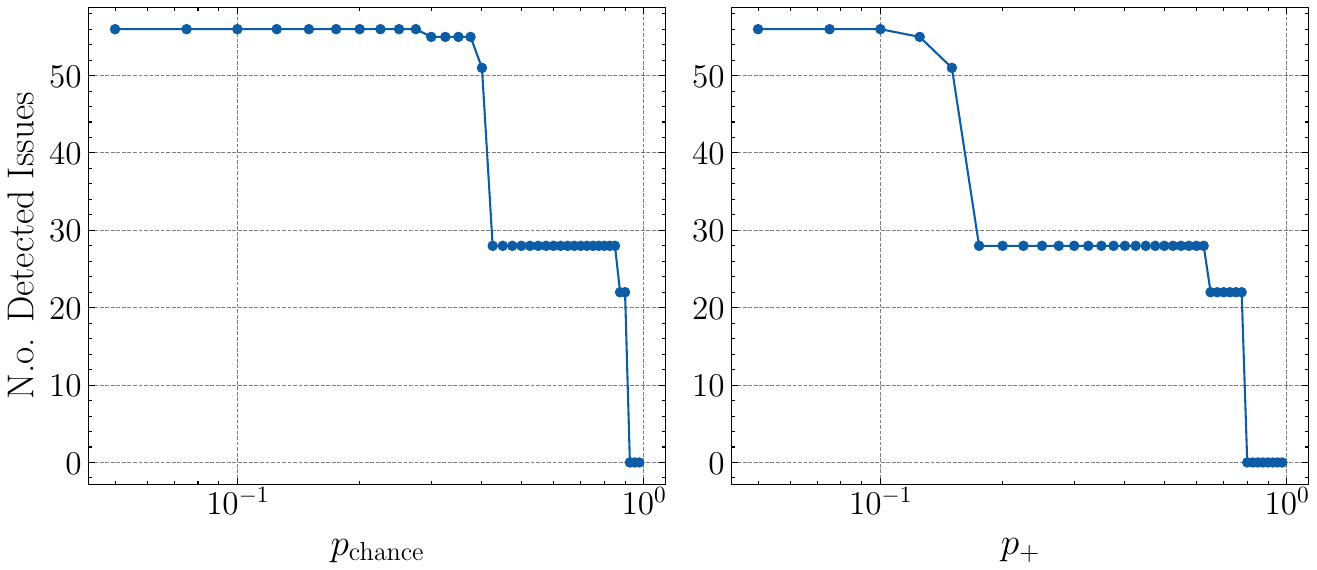}}

    \subfigure[Label Errors]{\label{fig:Stopping-Sensitivity-PAD-UFES-20-Labels}%
      \includegraphics[width=1.0\linewidth]{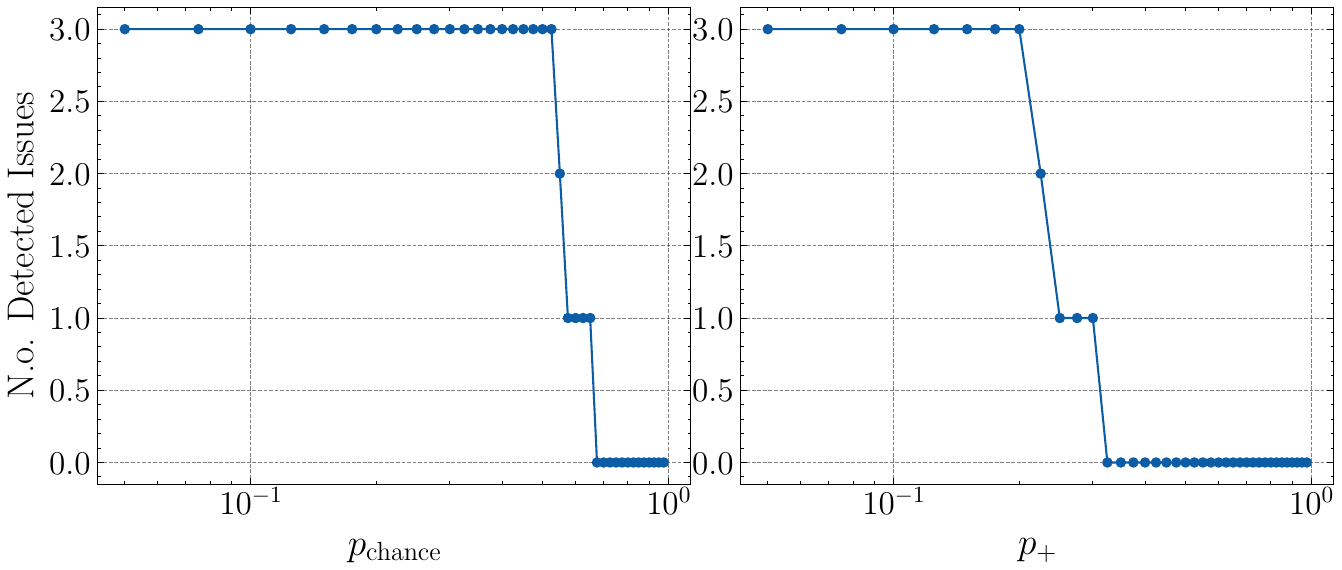}}
  }
\end{figure}

\newpage
\section{Details on human confirmation} \label{app:Human-Confirmation}
Annotators use the same annotation tool as in \citet{groger_selfclean_2023}, which is shown in Figure \ref{fig:Verification-Tool}. 
The annotation process starts with selecting a dataset and data quality issue (e.g. the DDI dataset and irrelevant samples) and then proceeds with binary questions (listed below) about single images or pairs thereof, depending on the task.
Note that the samples' ranks are not displayed to avoid potential bias.

\begin{description}
\item[Irrelevant samples:] 
    ``Your task is to judge if the image shown is irrelevant. 
    Select \textit{yes} when the image is not a valid input for the task at hand.''
        
\item[Near duplicates:]
    ``Your task is to judge whether the two images shown together are pictures of the same object. 
    Note that pictures of the same object can be identical or different shots with the same object of interest.''
    
\item[Label errors:]
    ``Your task is to judge whether the image's label is correct.
    Please select that the label is an error only if you think it is wrong and not when there is low uncertainty or ambiguity.''
\end{description}

\section{Ethics statement}
Our research does not require IRB approval because our study examines publicly available datasets and does not involve human subjects beyond annotations from co-authors.

Both medical experts and laypersons were not compensated financially but were instead acknowledged with co-authorship in a labeling consortium.

\begin{figure*}[htbp]
\floatconts
  {fig:Verification-Tool}
  {\caption{
        Screenshot of the verification tool from \citep{groger_selfclean_2023} used to confirm data quality issues.
  }}
  {%
    \includegraphics[width=1.0\textwidth]{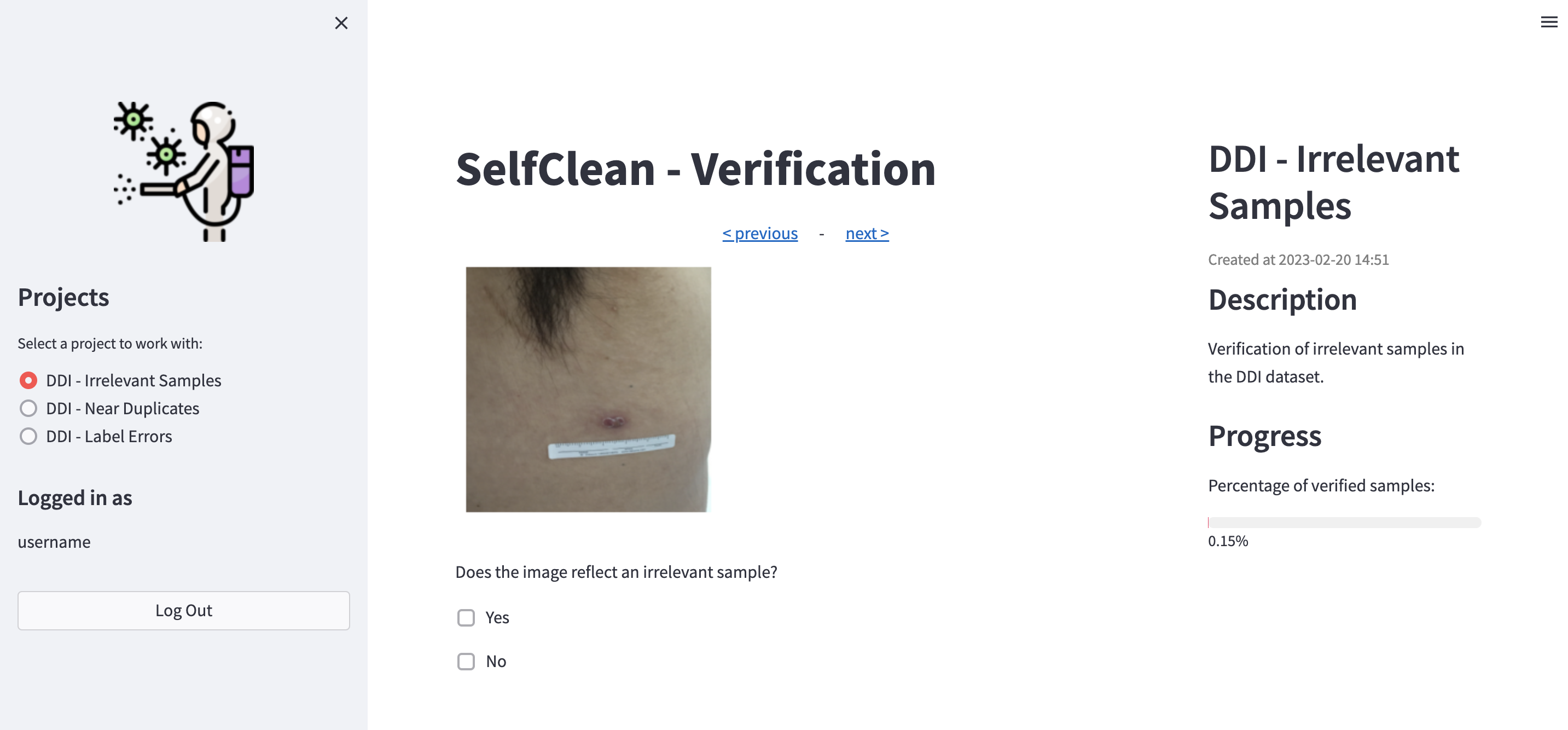}
  }
\end{figure*}

\end{document}